%% file: main.tex
\documentclass{article}

\PassOptionsToPackage{numbers}{natbib}

\usepackage[final]{neurips_2024}
\usepackage{graphicx}
\usepackage{xcolor}

\definecolor{brown}{RGB}{139, 90, 43}
\definecolor{teal}{RGB}{0, 128, 128}
\definecolor{plasmgreen}{HTML}{2E8B57}
\definecolor{carcinoyellow}{HTML}{B8860B}
\definecolor{granmagenta}{HTML}{C71585}

\usepackage{float}

\usepackage[utf8]{inputenc} 
\usepackage[T1]{fontenc}    
\usepackage{hyperref}       
\usepackage{cleveref}       
\usepackage{url}            
\usepackage{booktabs}       
\usepackage{enumitem}      
\usepackage{longtable}      
\usepackage{amsfonts}       
\usepackage{nicefrac}       
\usepackage{microtype}      
\usepackage{xcolor}         
\usepackage{graphicx}       
\usepackage{subcaption}     
\usepackage{multirow}
\usepackage{xcolor,colortbl}
\usepackage{adjustbox}
\usepackage{siunitx}
\usepackage[font=small]{caption}
\usepackage{placeins}
\usepackage{makecell}

\title{Atlas H\&E-TME: Scalable AI-Based Tissue Profiling at Expert Pathologist-Level Accuracy}

\author{
Kai Standvoss$^{*\;1}$, 
Miriam Hägele$^{*\;1}$, 
Rosemarie Krupar$^{*\;1}$, 
Julika Ribbat-Idel$^{*\;1}$, \AND
Jennifer Altschüler$^{1}$, 
Gerrit Erdmann$^{1}$, 
Hans Pinckaers$^{1}$, 
Evelyn Ramberger$^{1}$, 
Madleen Drinkwitz$^{1}$, \AND
Ádám Nárai$^{1}$, 
Alexander Möllers$^{1}$, 
Katja Lingelbach$^{1}$, \AND
Sebastian Kons$^{1}$, 
Lukas Hönig$^{1}$, 
Recepcan Adigüzel$^{1}$, 
Joana Baião$^{1}$, 
Alberto Megina Gonzalo$^{1}$, \AND
Marius Teodorescu$^{1}$, 
Marie-Lisa Eich$^{2\;3}$, 
Paolo Chetta$^{4}$, 
Shakil Merchant, \AND
Verena Aumiller$^{1}$,
Simon Schallenberg$^{2}$, 
Andrew Norgan$^{5}$, 
Klaus-Robert Müller$^{6\;7\;8\;9}$, \AND
Lukas Ruff$^{\dag\;1}$, 
Maximilian Alber$^{\dag\;1}$, 
Frederick Klauschen$^{\dag\;2\;5\;10\;11\;12}$ \\ \\
$^{1}$ Aignostics, Germany \AND 
$^{2}$ Institute of Pathology, Charité – Universitätsmedizin Berlin, Germany \AND 
$^{3}$ Berlin Institute of Health, Charité – Universitätsmedizin Berlin, Germany \AND 
$^{4}$ Dept.\ of Pathology, Massachusetts General Hospital, Harvard Medical School, Boston, MA, US \AND 
$^{5}$ Dept.\ of Laboratory Medicine and Pathology, Mayo Clinic, Rochester, MN, US \AND 
$^{6}$ Machine Learning Group, Technische Universität Berlin, Germany \AND 
$^{7}$ BIFOLD – Berlin Institute for the Foundations of Learning and Data, Germany \AND 
$^{8}$ Dept.\ of Artificial Intelligence, Korea University, Republic of Korea \AND 
$^{9}$ Max-Planck Institute for Informatics, Germany \AND 
$^{10}$ German Cancer Research Center (DKFZ) \& German Cancer Consortium (DKTK), \\ Berlin \& Munich Partner Sites, Germany \AND 
$^{11}$ Institute of Pathology, Ludwig-Maximilians-Universität München, Germany \AND 
$^{12}$ Bavarian Cancer Research Center (BZKF), Germany \AND
${*}$ Equal contribution \AND
$\dag$ Corresponding author}

\begin{document}

\maketitle

\begin{abstract}
Hematoxylin and eosin (H\&E) staining is the cornerstone of histopathology, yet scalable, quantitative analysis of H\&E whole-slide images (WSIs) remains a central challenge in computational pathology. 
We present Atlas H\&E-TME, an AI-based system built on the Atlas family of pathology foundation models that predicts tissue quality, tissue region, and cell type labels across multiple cancer types, yielding over 4,500 quantitative readouts per slide at cell-level resolution.\footnote{Atlas H\&E-TME v1.2.0 currently supports bladder, breast, colorectal, liver/biliary (hepatocellular carcinoma and cholangiocarcinoma), lung (non-small cell lung cancer and lung neuroendocrine tumors), pancreas, prostate, and stomach cancer types.} 
A key challenge to validating such systems is overcoming morphological ambiguity inherent to H\&E-only ground truth and the limited scalability of more informed references drawing on modalities such as immunohistochemistry (IHC). 
We address this with a dual validation framework combining biologically grounded depth with technical and morphological breadth. 
For depth, we propose an IHC-informed multi-pathologist consensus protocol that substantially improves inter-rater agreement over conventional H\&E-only annotation. 
This yields a molecularly grounded reference against which we compare Atlas H\&E-TME and pathologists working from H\&E alone. 
For breadth, we benchmark Atlas H\&E-TME on over 200,000 high-confidence H\&E-only pathologist annotations across 1,500+ cases spanning eight cancer types and their most common metastatic sites, with subtypes covering $>$90\% of clinical cases per cancer type, drawn from 25+ sources and 8+ scanner models. 
Benchmarked against the IHC-informed consensus, Atlas H\&E-TME matches or exceeds pathologist H\&E-only performance and generalizes consistently and robustly across this broad morphological and technical scope. 
In doing so, Atlas H\&E-TME turns the H\&E slide --- the most ubiquitous data in pathology --- into a scalable, quantitative window into the tumor and its microenvironment, laying a foundation for the next generation of tissue-based biomarkers in translational and clinical research.\footnote{Aignostics' \href{https://www.aignostics.com/products/atlas-he-tme/for-academics}{\textbf{Research Access Program}} offers academic researchers the ability to apply to run Atlas H\&E-TME on their own data. We applied Atlas H\&E-TME to TCGA and released it as the \href{https://huggingface.co/datasets/Aignostics/OpenTME}{\textbf{OpenTME}} dataset \cite{galama2026}.}
\end{abstract}

\section{Introduction}
\label{sec:intro}

Hematoxylin and eosin (H\&E) staining is the workhorse of histopathology, performed on essentially every tissue specimen examined for diagnostic or research purposes. 
The resulting H\&E whole-slide images (WSIs) carry rich information about tissue architecture and cellular composition that is central to characterizing tissue in general, and specifically tumors and their microenvironment (TME) --- the cellular ecosystem surrounding a tumor, whose composition and spatial organization are increasingly recognized as drivers of disease progression and therapeutic response \citep{fridman2017immune}. 
Quantitative characterization of tissue at scale has thus become a central goal of computational pathology, with direct applications in translational and clinical research, for example for biomarker discovery or patient stratification. 
Compared to multiplex protein assays such as immunohistochemistry (IHC) or multiplex immunofluorescence (mIF), which characterize cells through specific molecular markers but are either narrow in scope or more costly and difficult to scale, AI-based profiling of routine H\&E offers a route to comprehensive, cell-level tissue profiling across large cohorts without additional staining \citep{diao2021, park2022artificial, vanguri2024histotme}. 

We introduce Atlas H\&E-TME, an AI-based system for comprehensive spatial profiling of H\&E-stained WSIs at cell-level resolution. 
Built on the Atlas family of pathology foundation models \citep{dippel2024,alber2025,alber2026}, it composes tissue quality control, tissue compartment segmentation, and cell-type classification into over 4,500 quantitative readouts per slide, characterizing the principal stromal and immune cell populations across currently eight solid tumor types. 

Realizing this potential in practice, however, requires that such predictions are reliable across the morphological and technical diversity encountered in clinical and research practice --- and validating a system at this breadth and level of detail is itself challenging. 
Pathologist annotation of cell types on H\&E alone is the dominant approach to generating evaluation references, but is fundamentally limited in two ways. 
First, H\&E staining can be inherently insufficient for distinguishing morphologically similar cell types with certainty --- most pronounced for immune populations such as macrophages or plasma cells, whose nuclear morphology overlaps with that of other mononuclear cells on H\&E. 
Second, even where the morphology is informative, pathologist annotations on H\&E are subject to inter-rater variability \citep{kang2022variability, elmore2015diagnostic}, which can conflate model error with annotator disagreement and limit the reliability of any single-annotator H\&E reference. 
In response, more biologically grounded references can be constructed by informing annotation with additional modalities, e.g.~by coregistering IHC or mIF stains to the H\&E section \citep{li2025automated, ghahremani2022deepliif, azam2024samesection} or by restaining the same physical section next to H\&E imaging \citep{bulten2019epithelium,mrowiec2022immunohistochemistry,
koreuber2025phenobench}, but the cost of generating coregistered multi-marker data and consolidating multi-pathologist consensus annotations limits the practicality of generating large-scale references necessary to cover the broad morphological and technical scope of H\&E tissue profiling. 
Existing validation efforts have therefore typically focused on a single axis at a time: more informed references on small cohorts, or broader H\&E-only evaluation, often limited to certain indications. 

We address this tension through a dual validation approach. 
We validate Atlas H\&E-TME along two systematic axes:

\begin{itemize}[leftmargin=1.5em, itemsep=0.2em, topsep=0.3em]
  \item \textbf{In-depth (biologically grounded) validation} using an IHC-informed multi-pathologist consensus reference, constructed via a sequential bleach-and-restain workflow with a five-stain IHC panel and two-pass pathologist annotation (H\&E-only, then with coregistered IHC) on the same physical section. This protocol substantially improves inter-rater agreement over H\&E-only annotation and enables direct comparison of Atlas H\&E-TME against pathologists annotating from H\&E alone, on a molecularly grounded ground truth. 
  \item \textbf{In-breadth (generalizability) validation} on H\&E WSIs from more than 1{,}500 cases spanning eight different cancer types including primary tumors and metastases from their most common metastatic sites, with morphological subtypes covering $>$90\% of clinical cases per cancer type, drawn from 25+ sources and scanned on 8+ device types. More than 200,000 high-confidence H\&E-only pathologist annotations are gathered across this cohort ($>$100{,}000 each for cell classification and tissue segmentation), enabling systematic assessment of prediction consistency and robustness across the morphological and technical scope at which Atlas H\&E-TME is intended to operate.
\end{itemize}

Together, these two axes form a complementary validation framework. 
We argue that robust validation of H\&E-based AI systems requires this dual-axis design: depth to anchor performance against a biologically grounded reference, and breadth to ensure that performance is not an artifact of a narrow evaluation setting. 
The two protocols differ in scope: the in-depth comparison against pathologists focuses on cell classification, where H\&E ambiguity and inter-rater variability are most acute, while tissue quality control and tissue segmentation --- whose classes are defined by image-level and tissue architecture criteria --- are validated under the in-breadth protocol alone (Appendix~\ref{app:qc_ts_validation}). 

Our results show that the IHC-informed consensus establishes a meaningful, molecularly grounded reference, particularly on the morphologically ambiguous immune populations central to tissue immuno-profiling; that Atlas H\&E-TME matches or exceeds pathologist H\&E performance when both are referenced against this consensus; and that this performance generalizes consistently across the broad morphological and technical scope of the in-breadth evaluation.

\section{Related Work}
\label{sec:related_work}

\paragraph{H\&E-based tissue profiling for tumor and tumor microenvironment.}
Computational profiling of the tumor microenvironment from H\&E whole-slide images has become a central line of work in computational pathology, motivated by the prognostic and predictive value of tumor-infiltrating lymphocytes (TIL) \cite{salgado2015evaluation}, the spatial organization of stromal and immune compartments \cite{fridman2017immune}, and the relative cost and scalability advantages of H\&E over multiplex protein assays \cite{abousamra2021multi}. 
Early approaches focused on individual cell populations of clinical interest — most prominently TIL detection and density estimation on H\&E \cite{saltz2018, swiderska2019learning, acs2019open} — and on slide-level TIL scores that recover prognostic signals validated against pathologist consensus guidelines \cite{salgado2015evaluation, amgad2020report}. 
More recent efforts have moved beyond single-population scoring toward comprehensive characterization of the TME from routine H\&E. 
Commercial and research systems such as PathExplore \cite{diao2021}, Lunit SCOPE IO \cite{park2022artificial}, or HistoTME \cite{vanguri2024histotme} extract structured panels of human-interpretable features such as cell counts, densities, ratios, tissue compartment statistics, and spatial neighborhood features, that can serve as inputs to downstream biomarker discovery and patient stratification. 
Atlas H\&E-TME is positioned in this category of comprehensive H\&E tissue profiling systems, with broad indication coverage and a readout panel designed for translational and clinical research use, built on the family of Atlas pathology foundation models \cite{dippel2024,alber2025,alber2026}. 

\paragraph{Cell detection and classification models on H\&E.}
The cell detection and classification stage of H\&E tissue profiling has been subject of continuous methodological work. 
HoVer-Net \cite{graham2019} established a convolutional architecture for joint nuclear instance segmentation and classification, which was followed by extensions \cite{horst2024cellvit, baumann2024hovernext} as well as task-specific designs such as StarDist \cite{schmidt2018}. 
Recent work has shifted toward leveraging pathology foundation models as backbones for cell-level tasks. 
CellViT \cite{horst2024cellvit} and CellViT++ \cite{horst2025cellvitpp} use Vision Transformer encoders pretrained on large pathology corpora to obtain cell embeddings. 
HistoPLUS \cite{adjadj2025} combines a foundation model backbone with curated data for downstream cell classification training. 
In parallel, pathology foundation models themselves \cite{dippel2024, chen2024uni, vorontsov2024virchow, xu2024provgigapath, alber2025, alber2026} have demonstrated that self-supervised pre-training on large, diverse histopathology corpora yields representations that transfer effectively to downstream cell and tissue tasks. 
Atlas H\&E-TME builds on this line of work by composing tissue QC, tissue segmentation, and cell classification models on the Atlas foundation model backbone \cite{dippel2024,alber2025,alber2026}, with a cell taxonomy and indication scope that supports comprehensive tissue profiling readouts across solid tumor types. 

\paragraph{Validation strategies for H\&E cell classification.}
The dominant evaluation paradigm for H\&E cell classification has been pathologist-annotated benchmarks such as PanNuke \cite{gamper2019, gamper2020}, MoNuSAC \cite{verma2021monusac}, Lizard \cite{graham2021lizard}, and NuCLS \cite{amgad2022nucls}. 
These benchmarks have driven substantial methodological progress but share limitations relevant to the validation of tissue profiling systems: they are typically restricted in cancer type coverage, contain limited scanner and laboratory diversity, and rely on pathologist annotations from H\&E alone, which is known to be subject to inter-rater variability — especially for morphologically ambiguous immune populations such as macrophages, plasma cells, and granulocytes \cite{kang2022variability, elmore2015diagnostic}. 
A complementary line of work has sought to construct more informed ground truth by leveraging additional modalities: same-section or serial-section immunohistochemistry (IHC) and multiplex immunofluorescence (mIF) have been used to derive cell labels for training and validation, either by registering IHC/mIF to H\&E \cite{li2025automated,ghahremani2022deepliif,azam2024samesection} or by restaining the same physical section after H\&E imaging \cite{bulten2019epithelium,mrowiec2022immunohistochemistry}. 
PhenoBench \cite{koreuber2025phenobench} recently introduced mIF-derived granular cell phenotypes as an evaluation reference and showed that pathology foundation models that achieve macro F1 above 0.70 on previous H\&E-only benchmarks can drop as low as 0.20 on its more granular and technically diverse benchmark, providing evidence that earlier H\&E benchmarks may overstate generalization. 
Most existing efforts focus on a single axis at a time: either more biologically informed reference standards on specific cohorts, or broader H\&E-only evaluation, however often limited to a single or few indications or not accounting for inter-rater variability systematically. 
We propose a dual validation approach that combines both axes, using IHC-informed multi-pathologist consensus to establish a molecularly grounded reference for in-depth comparison, and a large-scale, stratified H\&E-only cohort spanning multiple cancer types, sites, scanners, and laboratories for the in-breadth assessment of consistency and robustness. 

\section{Atlas H\&E-TME}
\label{sec:heta}

Atlas H\&E-TME is an AI-based system for spatial tissue profiling of H\&E-stained WSIs at cell-level resolution. 
It is built on the Atlas family of pathology foundation models \cite{dippel2024,alber2025,alber2026}, co-developed by Aignostics, Charité – Universitätsmedizin Berlin, LMU Munich, and Mayo Clinic. 
The foundation model underlying its model components is a vision transformer (ViT) \cite{dosovitskiy2021} pre-trained on a large-scale, diverse histopathology dataset using self-supervised learning with the DINO framework \cite{oquab2023,simeoni2025}. 
Task-specific models were obtained by using the Atlas backbone in a supervised fashion on pathologist-informed annotations across diverse training data. 

Atlas H\&E-TME applies four stages to each WSI:

\paragraph{(1) Tissue Quality Control (QC)} segments the slide into regions of valid tissue, out-of-focus areas, tissue artifacts, pen marker regions, and no tissue background, ensuring that only tissue of sufficient quality is passed to downstream analysis. The QC model uses a multi-scale semantic segmentation architecture on top of the foundation model. Training employed a combined cross-entropy and Dice loss. 

\paragraph{(2) Tissue Segmentation} classifies valid tissue into seven distinct tissue types — carcinoma, (normal) epithelial tissue, stroma, necrosis, blood, vessel, and other — also using a multi-scale semantic segmentation architecture on top of the foundation model, providing both fine-grained tissue detail and broader tissue context. Training employed a combined cross-entropy and Dice loss. 

\paragraph{(3) Cell Detection \& Classification} identifies individual cells using a custom StarDist-based \cite{schmidt2018} nucleus segmentation model applied within segmented tissue regions, excluding blood and necrotic areas. Each detected cell is then classified into one of nine classes — carcinoma cell, (normal) epithelial cell, fibroblast, lymphocyte, plasma cell, macrophage, granulocyte, endothelial cell, and other — by a multi-scale classifier on top of the foundation model, providing both nuclear-level detail and surrounding tissue context. Training employed a focal loss to address class imbalance. 

\paragraph{(4) Tissue \& Cell Metrics} are computed based on all classified cells and tissue regions, further combined into neighborhood features. Readouts include cell counts, ratios, densities, and nuclear morphology features per cell type and are reported both at the slide level and stratified by tissue region type. Neighborhood analysis further quantifies spatial relationships between cell types by computing co-occurrence statistics, ratios, and densities within defined radii of 20\,\textmu m and 40\,\textmu m.

\begin{figure}[tbh]
\centering
\setlength{\fboxsep}{0em}

\begin{minipage}[t]{0.24\linewidth}
    \centering \sffamily
    \fbox{\includegraphics[width=0.95\linewidth]{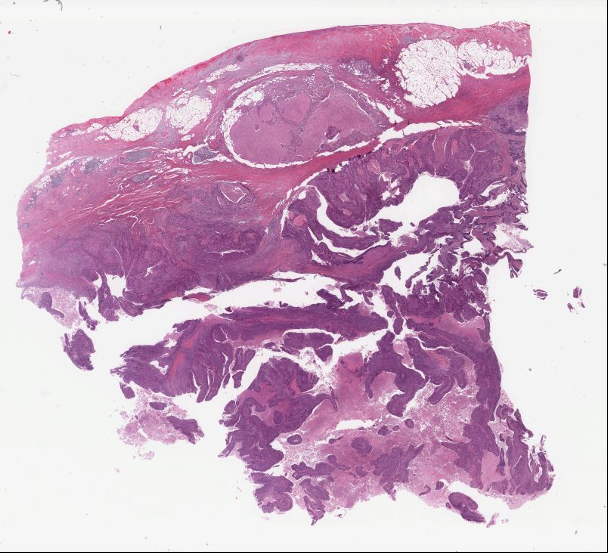}}\\[0.3em]
    \fbox{\includegraphics[width=0.95\linewidth]{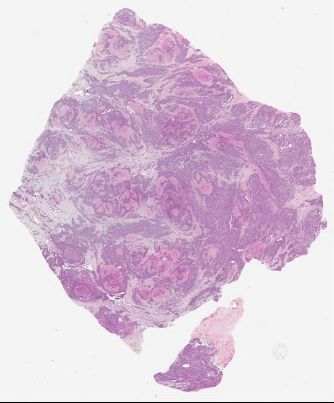}}\\[0.4em]
    WSI
\end{minipage}%
\begin{minipage}[t]{0.05\linewidth}
    \centering
    \hspace{0.99\linewidth}
\end{minipage}%
\begin{minipage}[t]{0.24\linewidth}
    \centering \sffamily
    \fbox{\includegraphics[width=0.95\linewidth]{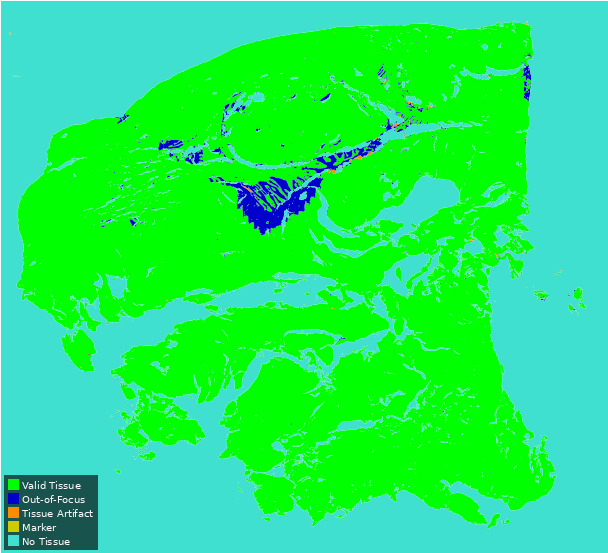}}\\[0.3em]
    \fbox{\includegraphics[width=0.95\linewidth]{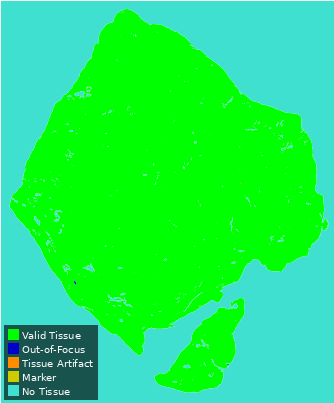}}\\[0.4em]
    Tissue QC
\end{minipage}%
\begin{minipage}[t]{0.24\linewidth}
    \centering \sffamily
    \fbox{\includegraphics[width=0.95\linewidth]{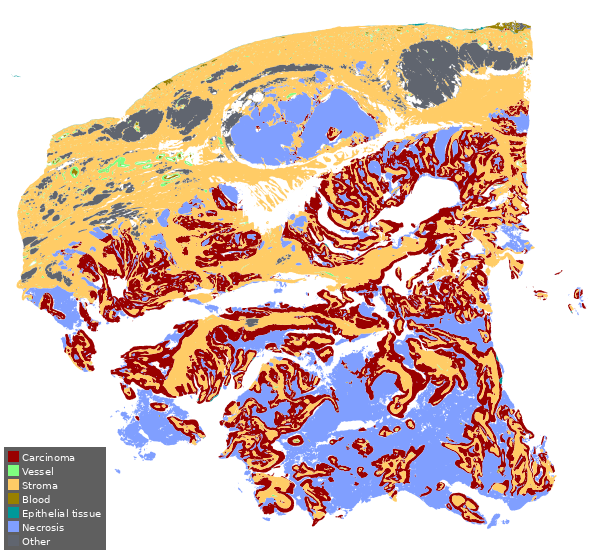}}\\[0.3em]
    \fbox{\includegraphics[width=0.95\linewidth]{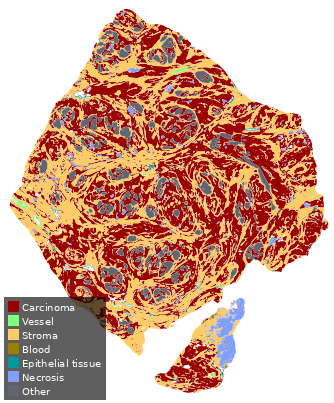}}\\[0.4em]
    Tissue Segmentation
\end{minipage}%
\begin{minipage}[t]{0.24\linewidth}
    \centering \sffamily
    \fbox{\includegraphics[width=0.95\linewidth]{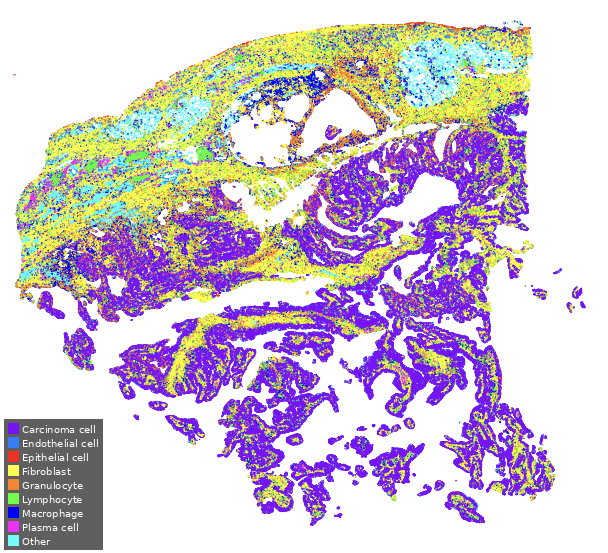}}\\[0.3em]
    \fbox{\includegraphics[width=0.95\linewidth]{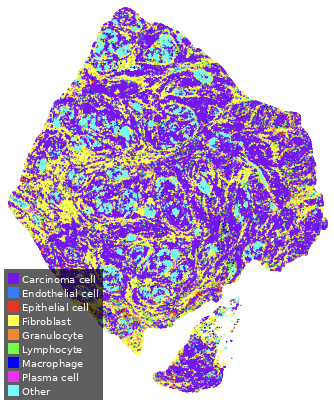}}\\[0.4em]
    Cell Classification
\end{minipage}
\caption{Two example WSIs from TCGA with Atlas H\&E-TME outputs across the Tissue QC, Tissue Segmentation, and Cell Classification stages. Model outputs are integrated to derive quantitative readouts per slide at cell-level resolution, including spatial neighborhood features. Examples drawn from the OpenTME dataset \cite{galama2026}.}
\label{fig:example_thumbnails}
\end{figure}

Examples of Atlas H\&E-TME model outputs on TCGA WSIs are shown in Figure~\ref{fig:example_thumbnails}. 
More detailed descriptions of the Tissue QC, Tissue Segmentation, and Cell Classification model classes are provided in Appendix~\ref{app:models}. 

The class taxonomies of Atlas H\&E-TME were defined together with board-certified pathologists to reflect the major tissue compartments and cell populations relevant to tissue profiling and characterizing the tumor microenvironment across solid tumor indications and will be further expanded in future versions. 
Tissue types were chosen to separate diagnostically and biologically distinct compartments (carcinoma, normal epithelium, stroma, necrosis, blood, vessels) while keeping the taxonomy general enough to apply across cancer types. 
Cell classes cover the principal stromal and immune cell populations — including lymphocytes, plasma cells, macrophages, granulocytes, fibroblasts, and endothelial cells — alongside carcinoma and normal epithelial cells, supporting downstream analyses of tissue-immune composition and spatial organization. 
The taxonomies and readouts were co-developed with the design of Atlas H\&E-TME's downstream applications in translational and clinical research, ensuring that model outputs map to biologically and clinically meaningful quantities. 

We validate Atlas H\&E-TME through a dual approach, combining IHC-informed in-depth evaluation with large-scale in-breadth benchmarking, as described in the next Section~\ref{sec:validation}. 

\section{A Dual Validation Approach: In-Depth and In-Breadth Evaluation}
\label{sec:validation}

As outlined in Section~\ref{sec:intro}, validating AI-based H\&E tissue profiling at the breadth and level of detail produced by Atlas H\&E-TME is challenging: H\&E-only pathologist references are limited by morphological ambiguity and inter-rater variability, while more informed IHC-based references are difficult to scale. 
We address this tension through a dual validation approach. 
In-depth validation uses a molecularly grounded pathologist consensus as ground truth and compares Atlas H\&E-TME directly against pathologists annotating from H\&E alone, trading scale for per-annotation informativeness. 
In-breadth validation uses high-confidence H\&E-only annotations from board-certified pathologists trained and tested on the Atlas H\&E-TME taxonomy (to minimize inter-rater variability through clear annotation guidelines), gathered across a broad scope of cancer types, sample types, sources, and scanners, trading per-annotation informativeness for the scale needed to assess consistency and robustness. 

We focus on the in-depth IHC-informed evaluation on the Cell Classification stage, where the cellular markers of the IHC panel directly resolve cell identity and where H\&E morphological ambiguity and inter-rater variability are therefore most consequential. 
Tissue QC and Tissue Segmentation classes are instead defined by image-level and tissue architectural criteria for which both limitations are substantially less pronounced; we accordingly validate these two models under the in-breadth protocol only and discuss this design choice, with the corresponding results, in Appendix~\ref{app:qc_ts_validation}.

\subsection{In-Depth Validation: IHC-Informed Consensus Evaluation}
\label{ssec:ihc_validation}

To benchmark Atlas H\&E-TME against a reference that mitigates both H\&E morphological ambiguity and inter-rater variability, we constructed an IHC-informed multi-pathologist consensus reference standard on a multi-indication resection cohort and used it to compare Atlas H\&E-TME against pathologists annotating from H\&E alone. 

\subsubsection{In-Depth Validation: Validation Setup}
\label{sssec:ihc_val_setup}

\begin{figure}[t]
  \centering
  \begin{subfigure}[t]{0.54\textwidth}
    \centering
    \begin{minipage}[c][5cm][c]{\linewidth}
      \centering
      \includegraphics[width=\linewidth,height=5cm,keepaspectratio]{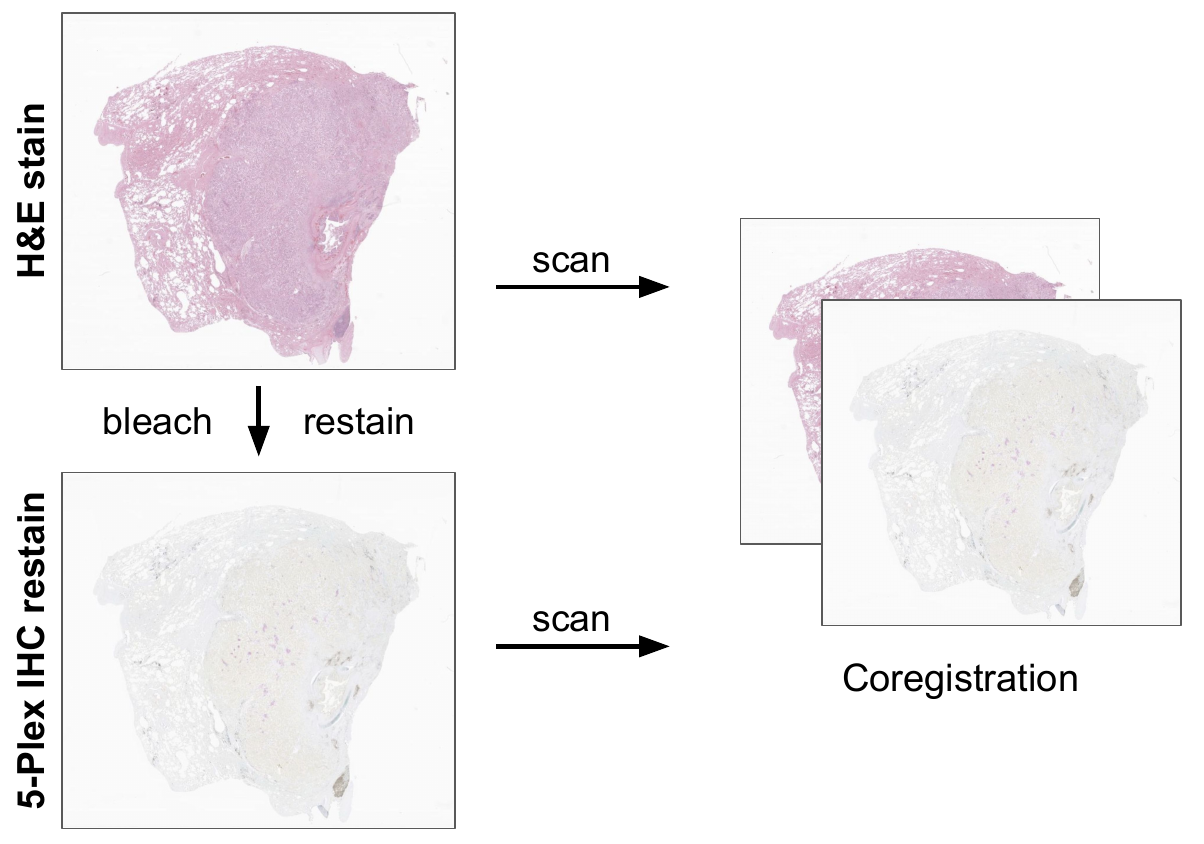}
    \end{minipage}
    \caption{H\&E bleach-and-restain with 5-plex IHC.}
    \label{fig:workflow-a}
  \end{subfigure}\hfill
  \begin{subfigure}[t]{0.44\textwidth}
    \centering
    \begin{minipage}[c][5cm][c]{\linewidth}
      \centering
      \includegraphics[width=\linewidth,height=4.25cm,keepaspectratio]{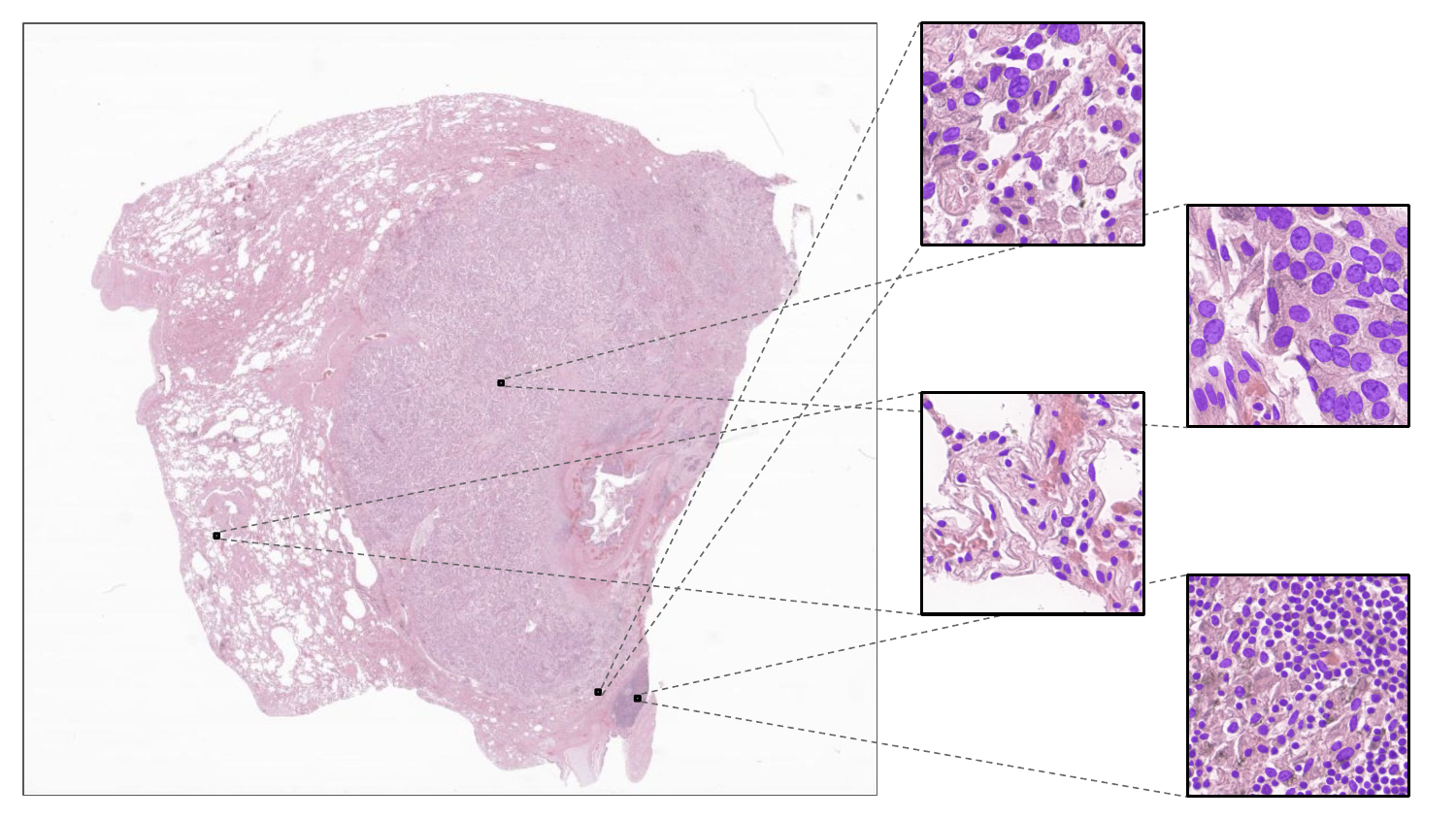}
    \end{minipage}
    \caption{Selection of 3--5 ROIs and cell detection.}
    \label{fig:workflow-b}
  \end{subfigure}
  
  \vspace{1em}

  \begin{subfigure}[t]{0.54\textwidth}
    \centering
    \begin{minipage}[c][5cm][c]{\linewidth}
      \centering
      \includegraphics[width=\linewidth,height=5cm,keepaspectratio]{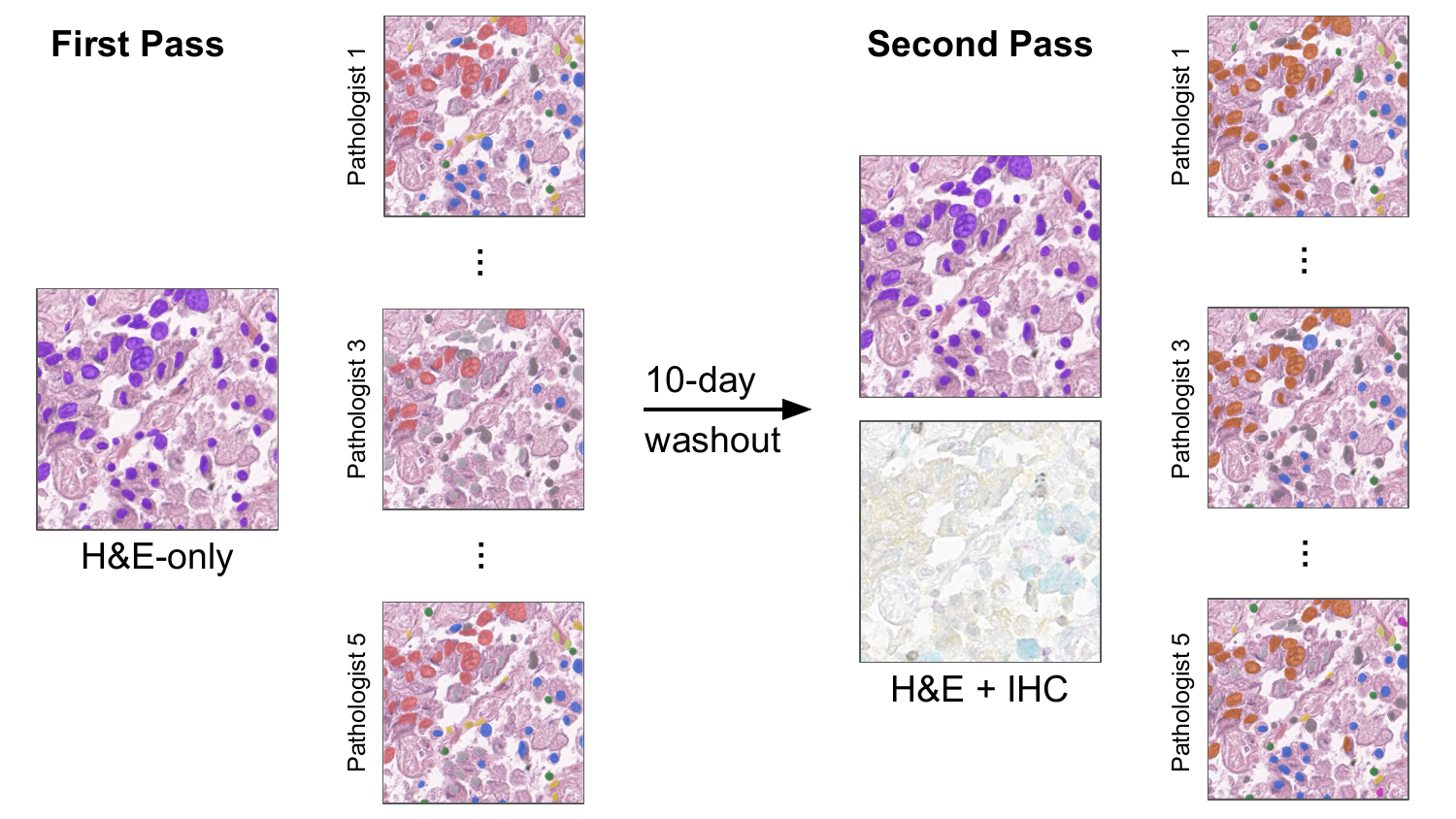}
    \end{minipage}
    \caption{Two-pass annotation with 10-day washout period.}
    \label{fig:workflow-c}
  \end{subfigure}\hfill
  \begin{subfigure}[t]{0.44\textwidth}
    \centering
    \begin{minipage}[c][5cm][c]{\linewidth}
      \centering
      \includegraphics[width=\linewidth,height=4.5cm,keepaspectratio]{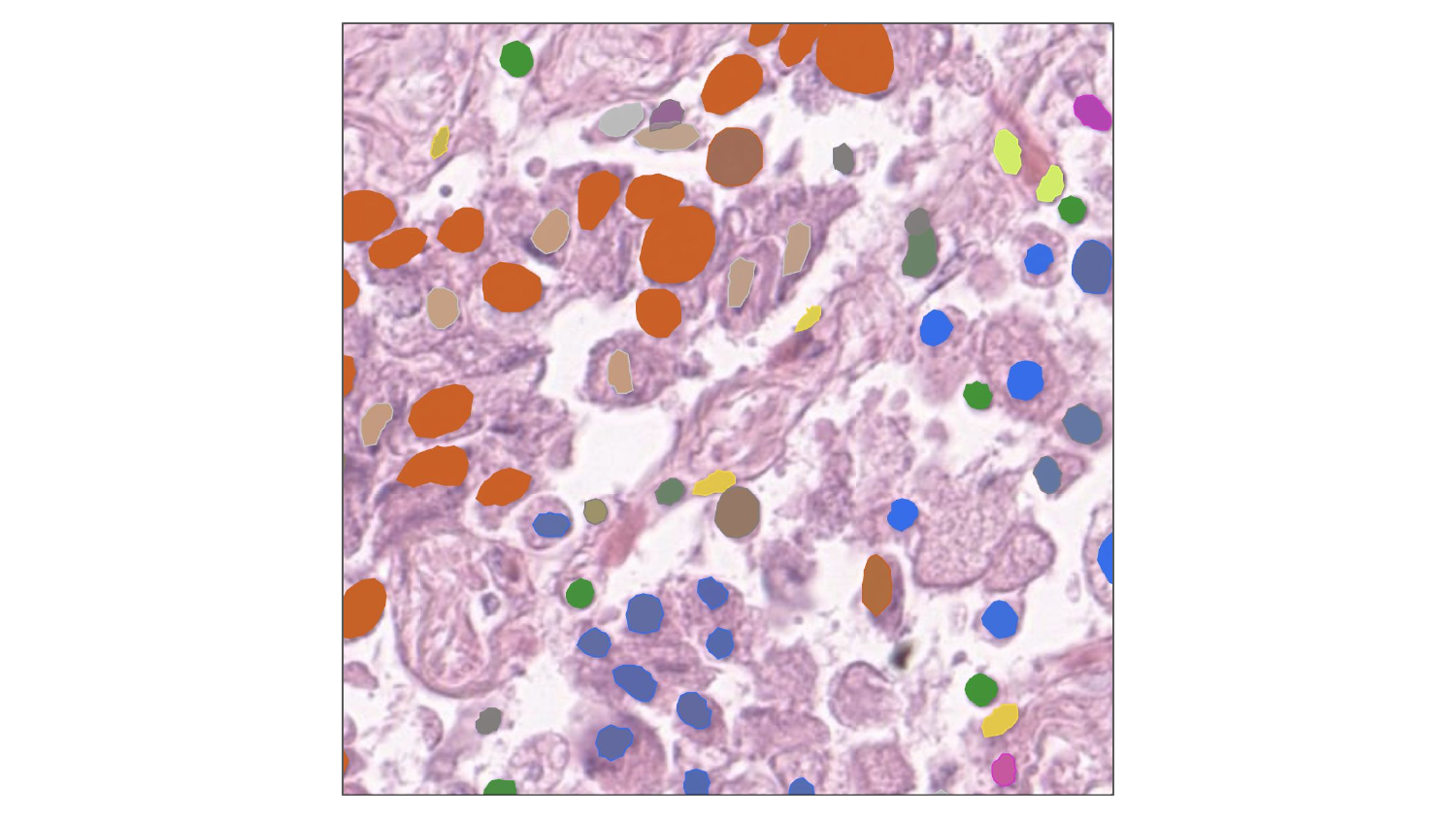}
    \end{minipage}
    \caption{IHC-informed 5-pathologist consensus.}
    \label{fig:workflow-d}
  \end{subfigure}
  \caption{%
    Overview of the in-depth validation workflow.
    \textbf{(\subref{fig:workflow-a})}~Each FFPE resection section is first H\&E-stained and scanned, then bleached and restained with a 5-plex IHC panel covering carcinoma cells, lymphocytes, granulocytes, plasma cells, and macrophages (cf.\ Table~\ref{tab:5plex_ihc_panel}); H\&E and IHC scans are coregistered to micrometer precision.
    \textbf{(\subref{fig:workflow-b})}~Within each WSI, 3--5 representative ROIs of 100$\times$100\,\textmu m are selected by pathologists to span tumor, tumor-stroma border, and stromal or benign compartments; cell instances are detected via the cell detection model (cf.\ Section~\ref{sec:heta}).
    \textbf{(\subref{fig:workflow-c})}~Five board-certified pathologists independently annotate each detected cell in two passes separated by a 10-day washout period: first on H\&E alone, then with the coregistered IHC overlay made available. 
    \textbf{(\subref{fig:workflow-d})}~An IHC-informed consensus reference is constructed by majority vote across the five pathologists' second-pass annotations.
  }
  \label{fig:ihc_val_workflow}
\end{figure}

\paragraph{Cohort and protocol.} The cohort comprises 30 whole-slide images from surgical resection specimens of different cases spanning three cancer indications — colorectal carcinoma, non-small cell lung cancer, and urothelial bladder carcinoma (10 cases per indication). 
Each section underwent a sequential bleach-and-restain workflow: H\&E staining and scanning were performed first, after which the same physical section was bleached and restained with a five-stain IHC panel covering the principal cell populations of the TME (Table~\ref{tab:5plex_ihc_panel}). 
H\&E and IHC scans were coregistered to micrometer precision. 
Within each WSI, 3--5 regions of interest (ROIs) of 100$\times$100\,\textmu m were selected by pathologists to span tumor, tumor-stroma border, and stromal or benign tissue compartments; regions dominated by necrosis or blood were avoided. 
Five board-certified pathologists then independently annotated each ROI in two passes separated by a ten-day washout period: a first pass on H\&E alone, followed by a second pass with the coregistered IHC overlay made available. 
To enable cell-level evaluation against model predictions, annotations were placed on cell instances detected by the StarDist-based \cite{schmidt2018} cell detection model (cf.~Section~\ref{sec:heta}), with pathologists assigning class labels to each detected cell. 
Figure~\ref{fig:ihc_val_workflow} provides an overview of the workflow. 

\begin{table}[h]
  \centering
  \small
  \caption{IHC markers in the 5-plex panel and their target cell classes. The dual CD68/MPO design enables disambiguation between macrophages and granulocytes within the same section.}
   \vspace{4pt}
  \label{tab:5plex_ihc_panel}
  \input{tab/5plex_markers}
\end{table}

\paragraph{Consensus construction.} The IHC-informed consensus reference was constructed by majority vote across the five pathologists' second-pass (IHC-informed) annotations per cell. 
Cells without a confident majority class --- those on which the pathologists disagreed or could not commit to a single class even with IHC --- were excluded from the evaluation ground truth. 
The resulting reference is both IHC-informed and consensus-based, addressing the two principal shortcomings of single-annotator H\&E ground truth — morphological ambiguity and inter-rater variability. 

\subsubsection{In-Depth Validation: Results}
\label{sssec:ihc_val_results}

\begin{figure}[tbh]
    \centering
    \includegraphics[width=0.85\textwidth]{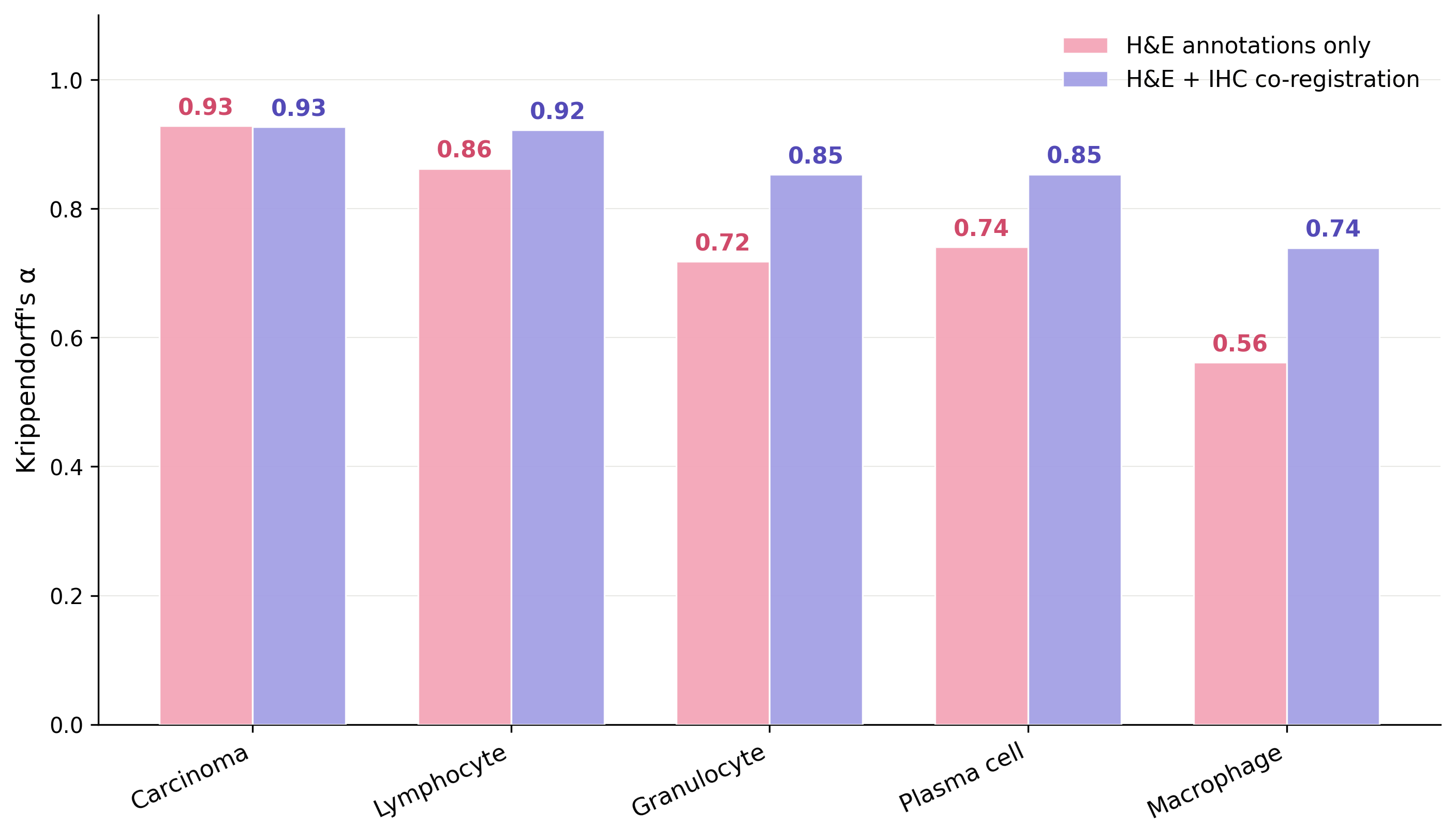}
    \caption{Inter-rater agreement (Krippendorff's $\alpha$) between the five pathologists across the five evaluated cell classes, comparing H\&E-only annotation to IHC-informed annotation. Agreement increases consistently from H\&E-only to IHC-informed annotation, with the largest gains on the morphologically most ambiguous cell classes. The result establishes the IHC-informed consensus as a more reliable ground truth.}
    \label{fig:5plex_alpha}
\end{figure}

\paragraph{IHC-informed consensus establishes a more reliable ground truth.} We first assessed whether the IHC-informed consensus is meaningfully more reliable than H\&E-only annotation by quantifying inter-rater agreement across the two annotation passes. 
Krippendorff's $\alpha$ \cite{krippendorff2018,zapf2016} between pathologists increases consistently from H\&E-only to IHC-informed annotation across all five evaluated cell classes (Figure~\ref{fig:5plex_alpha}), with the largest gains on the morphologically most ambiguous immune populations: granulocytes (0.72 → 0.85), plasma cells (0.74 → 0.85), and macrophages (0.56 → 0.74). 
Carcinoma cells, which are reliably identifiable from H\&E morphology alone, show the highest agreement, essentially unchanged from H\&E-only to IHC-informed annotation (0.93). 
The IHC-informed consensus thus establishes a more reliable reference precisely where H\&E-only ground truth is most uncertain and variable, justifying its use as the in-depth evaluation reference for the cell classes most relevant to TME profiling. 

\begin{figure}[tbh]
    \centering
    \includegraphics[width=0.85\textwidth]{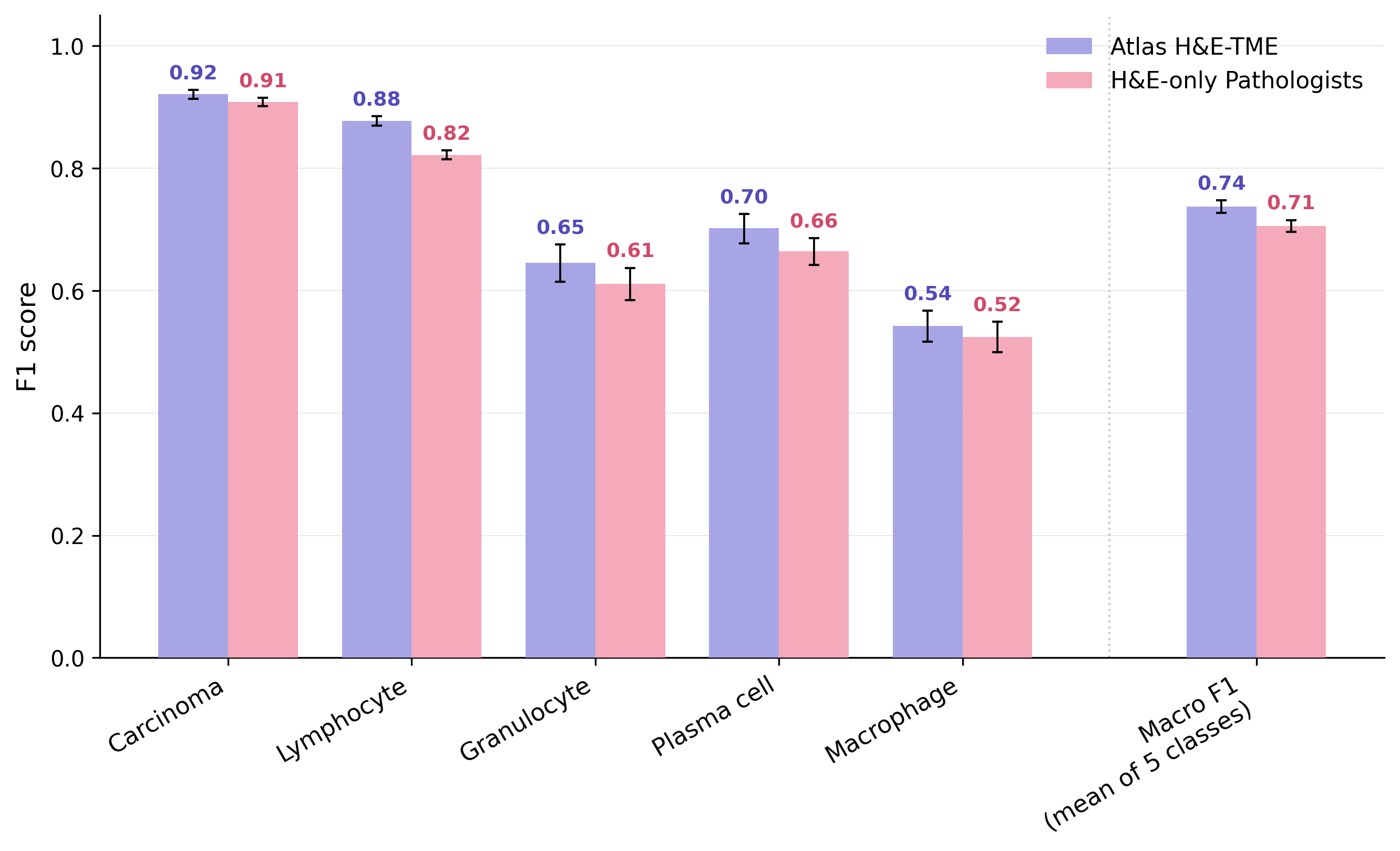}
    \caption{Per-class F1 score of Atlas H\&E-TME and the pathologist mean against the IHC-informed five-pathologist consensus. Atlas H\&E-TME matches or exceeds the mean pathologist H\&E-only performance across all five evaluated cell classes, with overlapping or favorable 95\,\% confidence intervals (10k stratified bootstrap iterations). Overall, Atlas H\&E-TME performs significantly better than the pathologist H\&E mean in Macro F1 (0.74 vs.\ 0.71; p = 0.0014, slide-clustered paired bootstrap).}
    \label{fig:5plex_f1}
\end{figure}

\paragraph{Atlas H\&E-TME matches or exceeds pathologist H\&E performance.} We next compared Atlas H\&E-TME against the five pathologists annotating from H\&E alone, with both evaluated against all cells of the IHC-informed consensus (Figure~\ref{fig:5plex_f1}). 
The results show that Atlas H\&E-TME matches or exceeds the pathologist mean F1 on all five evaluated cell classes, with overlapping or favorable 95\% confidence intervals throughout (10k stratified bootstrap iterations).
In three out of five classes (Carcinoma cell, Lymphocyte, Plasma cell), Atlas H\&E-TME scores better than every individual pathologist (Table~\ref{tab:in-depth_ranking}). 
While some per-class confidence intervals overlap, the macro-averaged F1 difference across all five classes is significant between Atlas H\&E-TME and pathologist H\&E mean (0.74 vs.\ 0.71; p = 0.0014, slide-clustered paired bootstrap). 
These results establish that Atlas H\&E-TME, predicting on H\&E alone, performs at the level of board-certified pathologists annotating on H\&E alone when both are referenced against the IHC-informed consensus — across the principal cell populations of the TME and across representative cancer indications. 

\begin{table}[h]
  \centering
  \small
  \caption{Per-class F1 score of each expert pathologist (P1--P5) and Atlas H\&E-TME. All scores are computed against the IHC-informed consensus reference. The last column shows Atlas H\&E-TME's rank among the six.}
  \vspace{4pt}
  \label{tab:in-depth_ranking}
  \input{tab/in-depth_ranking}
\end{table}

\begin{figure}[tbh]
    \centering
    \includegraphics[width=0.85\textwidth]{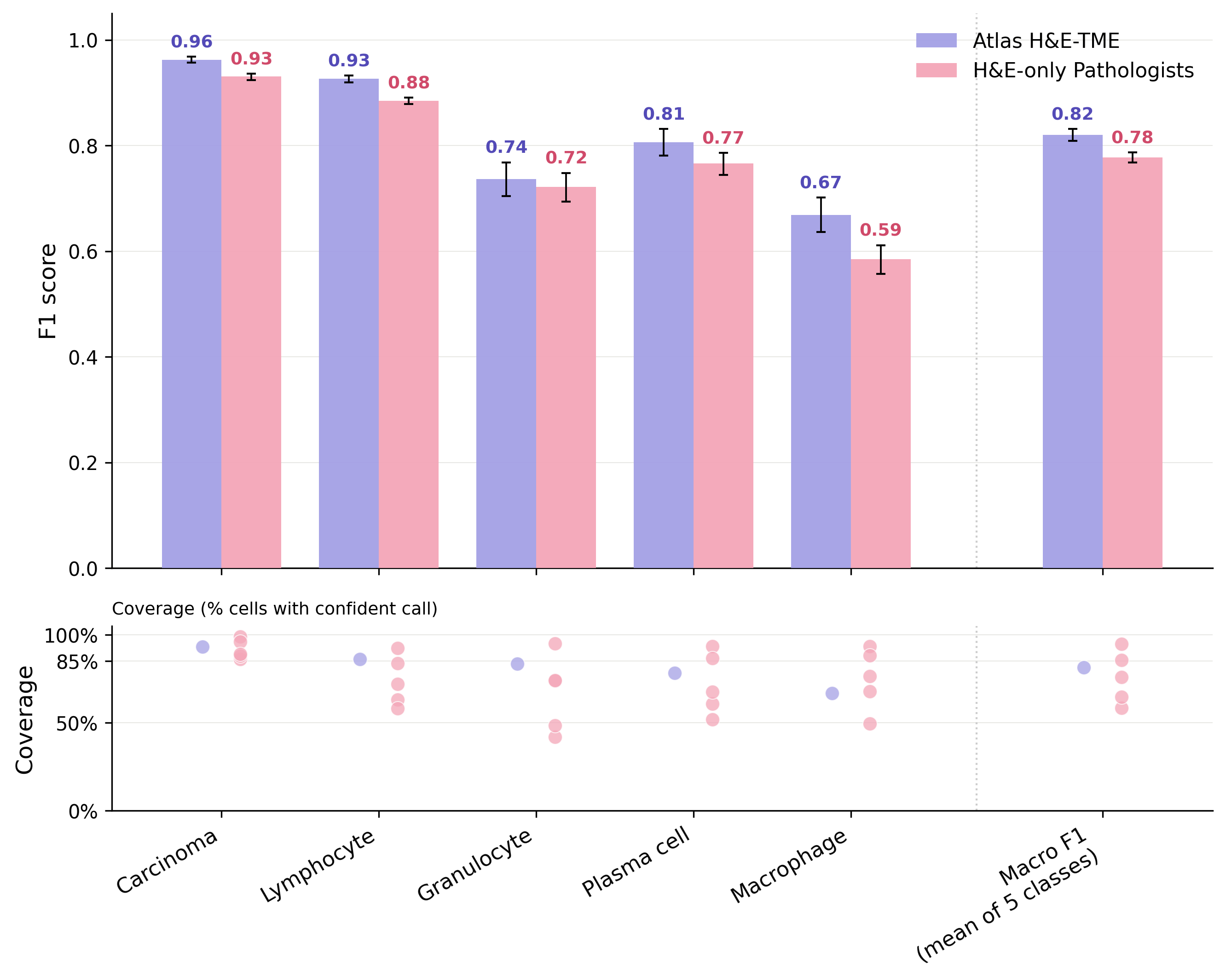}
    \caption{Per-class F1 score of Atlas H\&E-TME and the pathologist mean against the IHC-informed five-pathologist consensus when both may refrain from prediction: each pathologist is scored on cells they called with certainty and the model on the cells it retains after confidence thresholding at matched coverage. As expected, excluding ``uncertain'' cell calls raises absolute F1 for both over exhaustive calling (cf.\ Figure~\ref{fig:5plex_f1}). Atlas H\&E-TME continues to match or exceed the mean pathologist H\&E-only performance across all five evaluated cell classes (95\,\% confidence intervals from 10k stratified bootstrap iterations), indicating its confidence score is informative. Overall, Atlas H\&E-TME continues to significantly outperform the pathologist H\&E mean in Macro F1 (0.82 vs.\ 0.78; p < 0.0001, slide-clustered paired bootstrap). Coverage (bottom) for Atlas H\&E-TME is computed and matched over all detected cells, exceeding coverage on the five evaluated classes alone.}    
    \label{fig:5plex_f1_uncertain}
\end{figure}

\paragraph{Atlas H\&E-TME retains its advantage when both model and pathologists may refrain from prediction.} When a cell cannot be classified on H\&E with sufficient certainty, it can be preferable to abstain rather than force a call. 
We therefore repeated the comparison allowing both sides to refrain from prediction. On the H\&E-only pass, pathologists could decline to commit to a class by flagging a cell as ``uncertain'' --- the human analogue of a model abstaining. Each pathologist was then scored only on the cells they did not flag ``uncertain,'' and Atlas H\&E-TME only on the cells it retained after thresholding its prediction confidence, with the threshold set so the model's coverage matches the pathologists' mean coverage across classes. 
Each rater is thus evaluated on its own confidently called subset, making this a comparison at matched coverage (rather than on identical cells). 
As expected, absolute F1 scores rise for both model and pathologists relative to exhaustive calling (Figure~\ref{fig:5plex_f1_uncertain} vs.\ Figure~\ref{fig:5plex_f1}), showing that the deferred cells are also the most ambiguous. 
Atlas H\&E-TME continues to significantly outperform the pathologist H\&E mean in Macro F1 (0.82 vs. 0.78; p < 0.0001, slide-clustered paired bootstrap), and matches or exceeds the pathologist mean on all five classes, indicating that its confidence score is informative --- the cells it declines are disproportionately ``those it would most likely misclassify.''

We provide more details on the coverage-performance trade-off in Appendix~\ref{app:selective_prediction}, finding that Atlas H\&E-TME retains its advantage across the entire range of thresholds, not only at the matched coverage point.

We present results on another independent in-depth cohort with binary IHC stainings on NSCLC tissue microarrays (TMAs) in Appendix~\ref{app:capri}. 
Due to the single-marker (binary) IHC stainings on this TMA cohort, which limits the tissue context available per annotated cell constrained by binary H\&E–IHC coregistration, absolute values on this cohort are lower overall, but results confirm that Atlas H\&E-TME matches or exceeds pathologist H\&E-only performance.

\subsection{In-Breadth Validation: Large-Scale Evaluation on High-Confidence H\&E Annotations}
\label{ssec:he_validation}

While the in-depth evaluation in Section~\ref{ssec:ihc_validation} validates cell classification accuracy against a molecularly grounded reference, such validation protocol is limited in scale. 
Assessing how reliably this accuracy generalizes across the broad scope of Atlas H\&E-TME --- spanning multiple cancer types, primary and metastatic sites, sample types (biopsies, resections, and FNAs), scanners, and laboratories --- calls for a complementary validation protocol. 
In this section, we describe an in-breadth evaluation on a large, stratified cohort of confidently annotated H\&E whole-slide images, complementing the in-depth IHC-informed evaluation as the second arm of our dual validation approach. 

\subsubsection{In-Breadth Validation: Validation Setup}
\label{sssec:he_val_setup}

\paragraph{Cohort and scope.}
The in-breadth evaluation cohort consists of H\&E-stained FFPE WSIs from more than 1,500 cases spanning eight primary cancer types (Atlas H\&E-TME v1.2.0 scope) --- bladder, breast, colorectal, liver, lung, pancreas, prostate, and stomach --- as well as their five most common metastatic sites: bone, brain, liver, lung, and lymph node. 
Within each cancer type, slides cover the invasive morphological subtypes as defined by the WHO Classification of Tumours\footnote{https://tumourclassification.iarc.who.int} representing at least 90\% of clinical cases per cancer type, to capture the dominant axes of biological variation per indication. 

Primary and metastatic sites are evaluated as distinct evaluation entities, as for the same cancer type the surrounding host tissue exhibits different site-specific morphology. 
To support assessment of generalization across the conditions under which the model is intended to be used, slides span more than 25 laboratories and biobanks across Europe and the United States, and more than eight scanner models from the major vendors used in clinical and research pathology. 
The cohort is held out from model development and used exclusively for evaluation. 

\begin{figure}[t]
  \centering

  \begin{subfigure}[c]{0.58\textwidth}
    \centering
    \includegraphics[width=\linewidth]{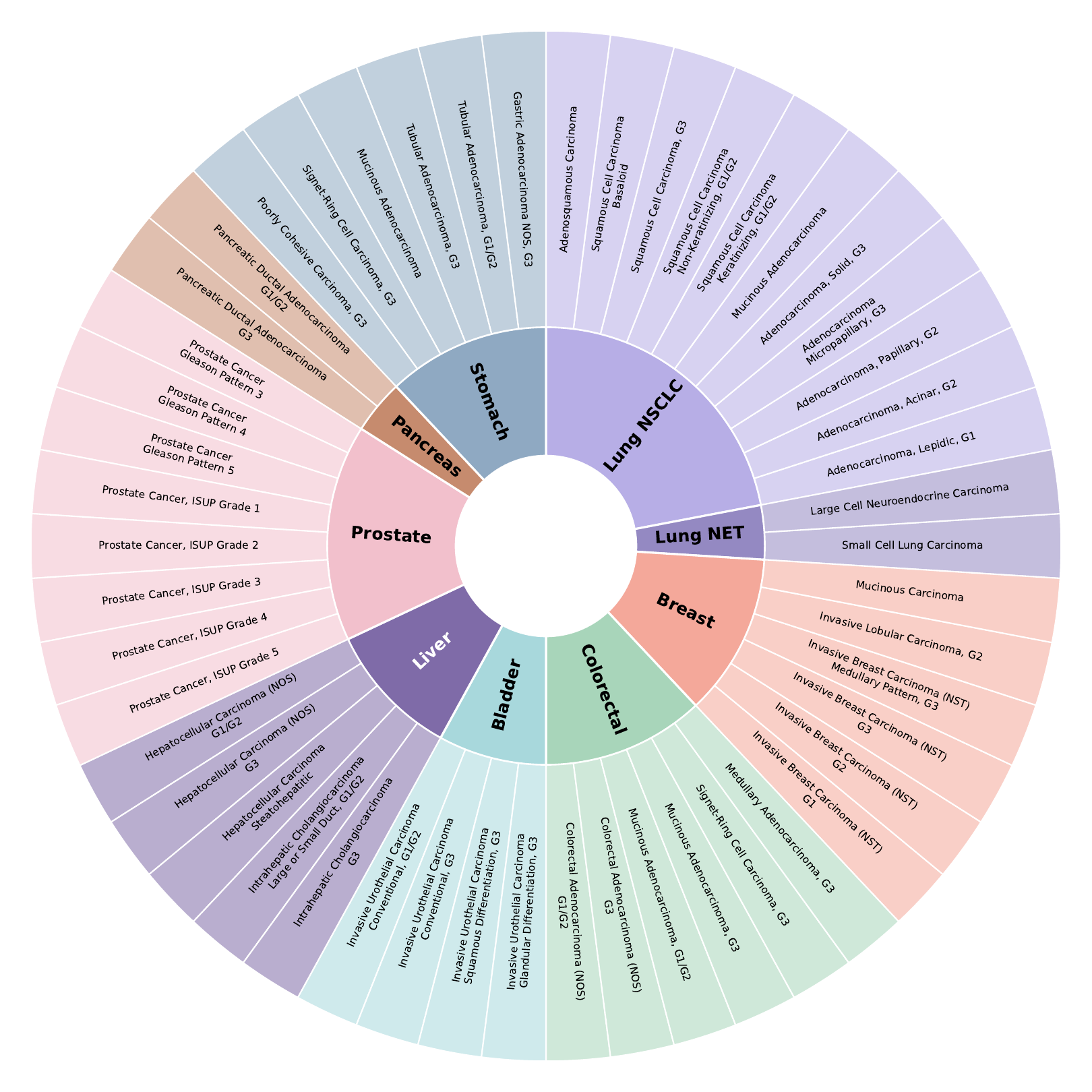}
    \caption{Morphological subtype coverage of primary cancer types.}
    \label{fig:he_val_primary}
  \end{subfigure}\hfill
  %
  \begin{subfigure}[c]{0.37\textwidth}
    \centering
    \begin{subfigure}[t]{\linewidth}
      \centering
      \includegraphics[width=0.85\linewidth]{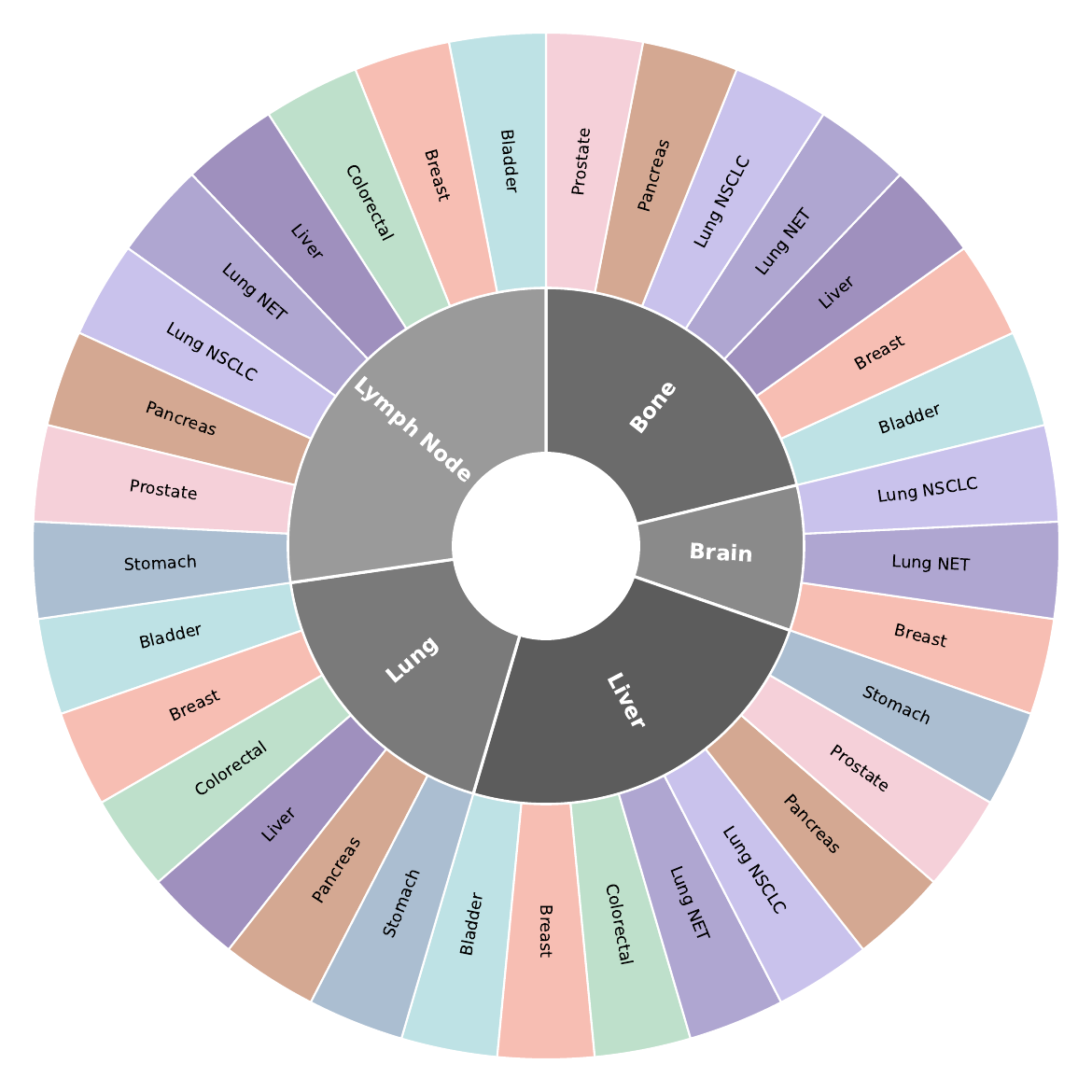}
      \caption{Metastatic site coverage.}
      \label{fig:he_val_metastases}
    \end{subfigure}

    \vspace{0.3em}

    \begin{subfigure}[t]{\linewidth}
      \centering
      \includegraphics[width=\linewidth]{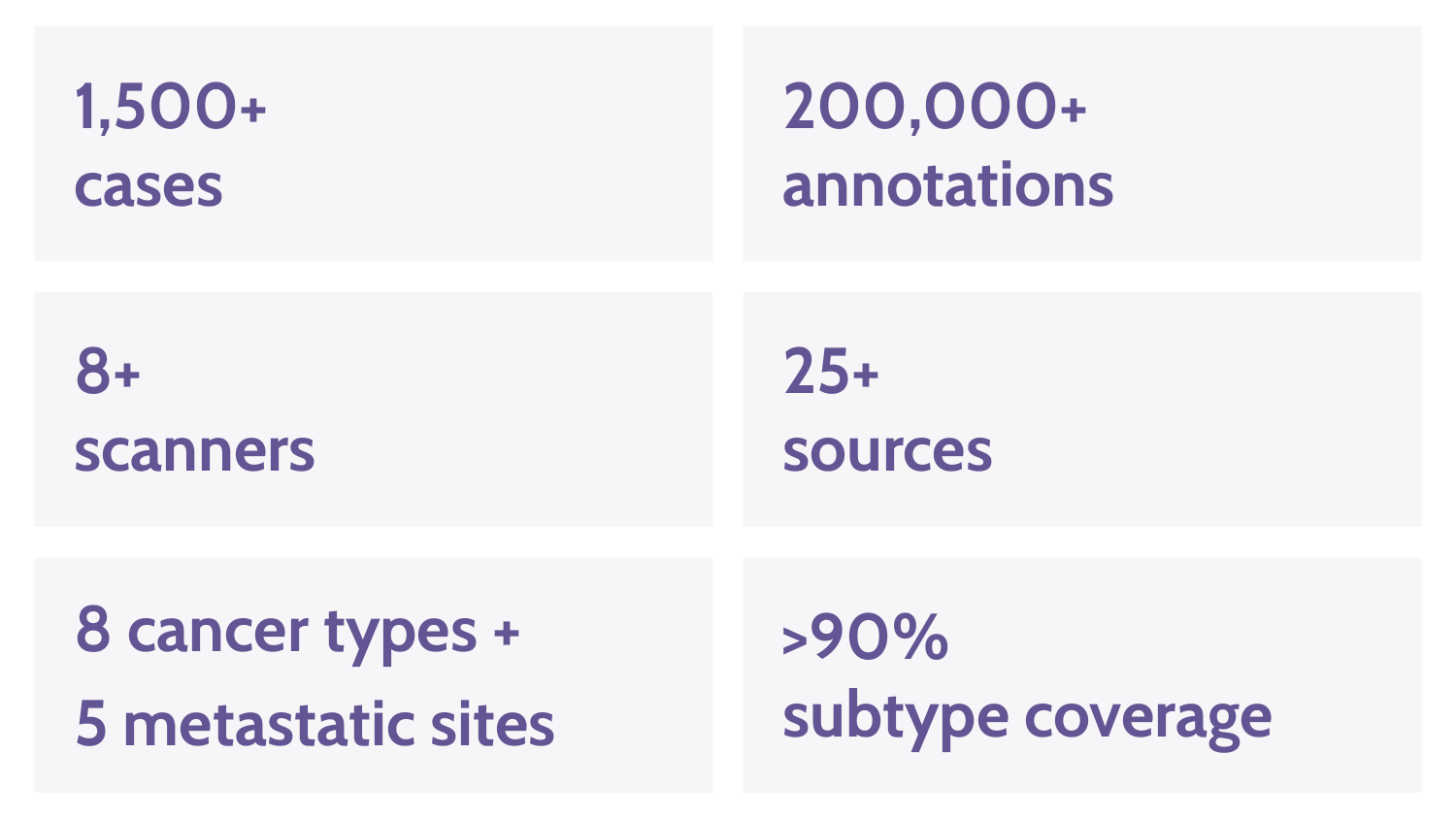}
      \caption{Cohort scale and diversity.}
      \label{fig:he_val_facts}
    \end{subfigure}
  \end{subfigure}
  
  \caption{%
    Scope of the in-breadth validation cohort.
    \textbf{(\subref{fig:he_val_primary})}~Morphological subtype coverage across the eight primary cancer types currently supported by Atlas H\&E-TME (Bladder, Breast, Colorectal, Liver, Lung, Pancreas, Prostate, Stomach). Each inner-ring wedge represents a cancer type; outer-ring wedges show the clinically relevant invasive morphological subtypes covered per cancer type, with $>$90\% coverage of subtypes per indication.
    \textbf{(\subref{fig:he_val_metastases})}~Primary cancer types evaluated at each of the five most common metastatic sites (bone, brain, liver, lung, lymph node), treated as distinct evaluation entities to account for site-specific host-tissue morphology.
    \textbf{(\subref{fig:he_val_facts})}~Key facts about size and diversity of the evaluation cohort for in-breadth validation.%
  }
  \label{fig:he_val_overview}
\end{figure}



\paragraph{Annotation protocol.}
Annotations on the in-breadth cohort were gathered by board-certified pathologists trained and tested on the Atlas H\&E-TME taxonomy (cf. Section~\ref{sec:heta}), with annotation guidelines specifying class definitions, morphological criteria, and annotation principles. 
For each WSI, on the order of one hundred cells were annotated across the slide, with annotation locations distributed representatively to cover morphological and staining variability. 
Annotations are balanced across cell classes to ensure each class is represented with sufficient sample size for evaluation, with carcinoma cells --- defining the key variability across morphological subtypes --- importantly representing approximately half of the annotations on primary-site slides and at least one third of annotations on metastatic slides to enable carcinoma subtype-specific assessment. 
In contrast to the in-depth evaluation, annotators worked exclusively from H\&E without access to IHC, enabling broad scaling of evaluation. 
To mitigate inter-rater variability at scale, annotators were trained and instructed to only label cells they could confidently classify; cells they could not confidently assign on H\&E alone were left unannotated rather than labeled without confidence. 
Each cell was annotated by a single pathologist, with annotations peer-reviewed by the board-certified pathologists who defined the taxonomy and led annotator training to ensure consistent annotation with sufficient quality. 
Pathologists working from H\&E alone thereby serve as the reference standard for the in-breadth evaluation, while some residual label noise from annotator overconfidence may nonetheless remain, particularly for morphologically ambiguous classes.  
In total, more than 100,000 high-confidence H\&E-only pathologist cell annotations representative for the broad scope of Atlas H\&E-TME were gathered. 

\subsubsection{In-Breadth Validation: Results}
\label{sssec:he_val_results}

Atlas H\&E-TME cell classification predictions are evaluated against pathologist annotations across the eight primary cancer types using the F1 score, reported separately for primary sites (Table~\ref{tab:in-breadth_primary}) and metastatic sites (Table~\ref{tab:in-breadth_metastatic}). 
Macro F1 is computed as the unweighted mean of per-class F1 scores. 

\begin{table}[h]
  \centering
  \small
  \setlength{\tabcolsep}{4pt}
  \caption{Atlas H\&E-TME in-breadth prediction performance (F1 Score) on primary sites across cancer types for all cell types. Results are averaged across annotations over WSIs from diverse laboratories and scanners.} 
   \vspace{4pt}
  \label{tab:in-breadth_primary}
  \input{tab/in-breadth_primary}
\end{table}

\begin{table}[h]
  \centering
  \small
  \caption{Atlas H\&E-TME in-breadth prediction performance (F1 Score) on metastatic sites of included primary cancer types for all cell types. Results are averaged across annotations over WSIs from diverse laboratories and scanners.}
   \vspace{4pt}
  \label{tab:in-breadth_metastatic}
  \input{tab/in-breadth_metastatic}
\end{table}

\paragraph{Atlas H\&E-TME performs robustly and consistently across cancer types.} Atlas H\&E-TME attains macro F1 cell classification scores across the eight covered cancer types on primary sites and metastatic sites between 0.85 and 0.95, with 11 out of 14 evaluation entities with macro F1 scores of at least 0.9. 
Notably, carcinoma cell classification F1 scores are consistently $\geq$ 0.96 at every site. 
As expected, macrophages and plasma cells show most variable numbers, reflecting inherent morphological complexity and thus ambiguity. 
Largest deviations are observed for macrophages and plasma cells at bone and liver metastatic sites, reflecting the morphological overlap with resident phagocytes in marrow and hepatic sinusoidal contexts. 
Overall, the in-breadth evaluation demonstrates that performance is robust and consistently close to pathologist-level across a large morphological and technical scope. 

\section{Conclusion}
\label{sec:conclusion}

We introduced Atlas H\&E-TME, an AI-based system for comprehensive spatial profiling of H\&E-stained whole-slide images that composes tissue quality control, tissue segmentation, and cell classification on the Atlas family of pathology foundation models to produce over 4,500 quantitative readouts per slide at cell-level resolution across eight solid tumor types, supporting downstream spatial analysis of tissue and the tumor microenvironment. 
Our central methodological contribution is a dual validation approach that pairs an IHC-informed multi-pathologist consensus reference for in-depth evaluation with a large-scale, stratified H\&E-only benchmark for in-breadth evaluation. 
This design addresses the two principal shortcomings of H\&E-only ground truth --- morphological ambiguity and inter-rater variability --- establishing a rigorous validation protocol. 

Across both validation axes, the results demonstrate Atlas H\&E-TME's reliability, consistency, and robustness for H\&E tissue profiling at scale. 
The IHC-informed consensus substantially improves inter-rater agreement over H\&E-only annotation, most markedly on the morphologically ambiguous immune populations central to the TME. 
Referenced against this more reliable standard, Atlas H\&E-TME's cell classification performs at the level of board-certified pathologists. 
In breadth, cell classification is robust and consistent across eight cancer types, spanning primary and metastatic evaluation entities, with macro F1 between 0.85 and 0.95 ($11$ of $14$ entities $\geq 0.9$) and carcinoma classification $\geq$ 0.96 at every site. 
The Tissue QC and Tissue Segmentation models generalize comparably across the same scope. 
The two axes are mutually reinforcing: depth anchors accuracy against a molecularly grounded reference, and breadth establishes that this accuracy holds across the morphological and technical diversity of translational and clinical research. 

Our evaluation also delineates where H\&E profiling is challenging. 
Performance is most variable for macrophages and plasma cells, with the lowest scores at the bone and liver metastatic sites, where they are most easily confused with morphologically similar resident cells. 
These deviations reflect a limitation intrinsic to H\&E rather than a specific model deficiency --- the same ambiguity that motivated the IHC-informed in-depth protocol, and that an H\&E-only reference cannot, by construction, fully resolve for these classes. 
Our results nevertheless show that strong performance of such challenging classes is achievable at scale, despite limitations inherent to H\&E. 
Responsible use requires understanding these limitations, both for drawing sound conclusions and for trustworthy deployment. 
Our benchmarking, while extensive, likewise remains finite, and its continued deepening and broadening stays a critical priority as the system's scope grows. 

Several directions follow from this work. 
On the validation side, we plan to extend in-depth validation with additional cohorts and broaden in-breadth validation alongside the growing scope of Atlas H\&E-TME. 
Atlas H\&E-TME remains under continued development, with ongoing work expanding cancer type coverage beyond the current eight indications and enriching the readout panel with additional cell- and tissue-level features. 
We applied Atlas H\&E-TME to TCGA and released it as the \href{https://huggingface.co/datasets/Aignostics/OpenTME}{OpenTME} dataset \cite{galama2026}, which we will expand with future cancer type coverage. 
Through Aignostics' \href{https://www.aignostics.com/products/atlas-he-tme/for-academics}{Research Access Program}, academic researchers can apply to run Atlas H\&E-TME on their own data, and we welcome feedback and feature requests from the research community. 

\section*{Acknowledgments}
\label{sec:acks}
We would like to thank the teams at Aignostics, the Department of Pathology at Charité – Universitätsmedizin Berlin, and the Department of Pathology at LMU Munich for their contributions to the development and validation of Atlas H\&E-TME. 
We particularly thank Aignostics' ML Engineering and Backend teams for their support in building and operating the underlying infrastructure. 

\bibliographystyle{plain} 
\bibliography{references} 

\newpage
\appendix

\section{Atlas H\&E-TME: Description of Model Classes}
\label{app:models}

The following Tables~\ref{tab:tissue_qc_classes}, \ref{tab:tissue_segmentation_classes}, and \ref{tab:cell_classification_classes} describe the classes of the Tissue QC, Tissue Segmentation, and Cell Classification model, respectively, in more detail.

\begin{table}[tbh]
  \centering
  \caption{Description of Tissue Quality Control model classes.}
   \vspace{4pt}
  \label{tab:tissue_qc_classes}
  \input{tab/classes_tissue_qc}
\end{table}

\begin{table}[tbh]
  \centering
  \caption{Description of Tissue Segmentation model classes.}
   \vspace{4pt}
  \label{tab:tissue_segmentation_classes}
  \input{tab/classes_tissue_segmentation}
\end{table}

\begin{table}[tbh]
  \centering
  \caption{Description of Cell Classification model classes.}
   \vspace{4pt}
  \label{tab:cell_classification_classes}
  \input{tab/classes_cell_classification}
\end{table}

\clearpage

\section{Validation of Tissue QC and Tissue Segmentation}
\label{app:qc_ts_validation}

The Tissue QC and Tissue Segmentation models of Atlas H\&E-TME are validated under the in-breadth protocol only, on the cohort of confidently annotated H\&E WSIs introduced in Section~\ref{ssec:he_validation}. 
We forego the IHC-informed in-depth protocol for these two models because the two principal limitations it addresses for cell classification --- H\&E morphological ambiguity at cell-level resolution and inter-rater variability on the most ambiguous immune populations --- do not directly transfer to the tissue compartment and QC level. 
While morphological ambiguity and inter-rater variability are not absent at this level, tending to concentrate at transition zones between compartments, they are generally less pronounced on a regional, tissue-architecture level than at cell-level resolution, and a cell-level IHC panel would not directly resolve them. 
Moreover, the bulk of Tissue QC classes are defined by non-biological artifacts (e.g., pen markers or out-of-focus regions). 

\paragraph{Annotation protocol.} Analogous to the cell classification in-breadth validation annotation protocol described in Section~\ref{sssec:he_val_setup}, we consider high-confidence H\&E-only annotations for the validation of the Tissue QC and Tissue Segmentation models. 
Both models are evaluated on the same held-out in-breadth cohort (Section~\ref{sssec:he_val_setup}), using region-level annotations rather than per-cell labels: annotators delineate contiguous regions and assign fine-grained, organ-specific tissue and artifact categories that are fused into the reported model classes. 
Regions are placed to densely sample compartment and artifact boundaries --- where class confusions concentrate --- and complemented by annotations distributed across the slide to capture morphological and staining variability within each class. 
As for cell classification, Tissue Segmentation annotation emphasizes carcinoma to enable subtype-specific assessment: approximately half of primary-site annotations fall on carcinoma, and on metastatic slides at least one third fall on carcinoma and a further third on host tissue to assess benign-tissue performance at the metastatic site. 
Tissue Segmentation regions were annotated by board-certified pathologists, reflecting the architectural reference standard for tissue compartments, whereas Tissue QC regions --- defined by image-level properties without biological correlates --- were annotated by trained annotators; as Tissue QC is independent of disease-subtype morphology, it was annotated on a representative subset of slides per primary site. 
All annotators were trained and qualified under a standardized procedure, with annotations again peer-reviewed. 
In total, more than 100,000 region-level Tissue Segmentation annotations and approximately 10,000 Tissue QC annotations were gathered across the in-breadth cohort. 

\paragraph{Results.}
Tables~\ref{tab:in-breadth_primary_ts}, \ref{tab:in-breadth_metastatic_ts}, \ref{tab:in-breadth_primary_qc}, and \ref{tab:in-breadth_metastatic_qc} report Tissue Segmentation and Tissue QC F1 scores on primary and metastatic sites, mirroring the cell classification reporting of Tables~\ref{tab:in-breadth_primary} and \ref{tab:in-breadth_metastatic}. 
Tissue Segmentation macro F1 ranges from 0.92 to 0.96 across primary sites and from 0.92 to 0.95 across metastatic sites, with carcinoma compartment F1 consistently $\geq$ 0.94 across every evaluated entity. 
Tissue QC macro F1 ranges from 0.89 to 0.96 across primary sites and from 0.92 to 0.98 across metastatic sites, with valid-tissue and no-tissue background F1 consistently $\geq$ 0.96. 
As in the cell classification reporting (Table~\ref{tab:in-breadth_metastatic}), F1 entries marked ``--'' denote classes that are not represented in the evaluation slice of the corresponding entity (e.g., normal epithelial tissue at bone or brain metastatic sites). 

\begin{table}[h]
  \centering
  \small
  \setlength{\tabcolsep}{4pt}
  \caption{Atlas H\&E-TME in-breadth Tissue Segmentation performance (F1 Score) on primary sites across cancer types for all tissue compartments. Results are averaged across annotations over WSIs from diverse laboratories and scanners.}
   \vspace{4pt}
  \label{tab:in-breadth_primary_ts}
  \input{tab/in-breadth_primary_ts}
\end{table}

\begin{table}[h]
  \centering
  \small
  \caption{Atlas H\&E-TME in-breadth Tissue Segmentation performance (F1 Score) on metastatic sites of included primary cancer types for all tissue compartments. Results are averaged across annotations over WSIs from diverse laboratories and scanners.}
   \vspace{4pt}
  \label{tab:in-breadth_metastatic_ts}
  \input{tab/in-breadth_metastatic_ts}
\end{table}

\begin{table}[h]
  \centering
  \small
  \setlength{\tabcolsep}{4pt}
  \caption{Atlas H\&E-TME in-breadth Tissue QC performance (F1 Score) on primary sites across cancer types for all QC classes. Results are averaged across annotations over WSIs from diverse laboratories and scanners.}
  \vspace{4pt}
  \label{tab:in-breadth_primary_qc}
  \input{tab/in-breadth_primary_qc}
\end{table}

\begin{table}[h]
  \centering
  \small
  \caption{Atlas H\&E-TME in-breadth Tissue QC performance (F1 Score) on metastatic sites of included primary cancer types for all QC classes. Results are averaged across annotations over WSIs from diverse laboratories and scanners.}
   \vspace{4pt}
  \label{tab:in-breadth_metastatic_qc}
  \input{tab/in-breadth_metastatic_qc}
\end{table}

\clearpage

\section{Uncertainty and Selective Prediction}
\label{app:selective_prediction}
 
In the in-depth validation (\Cref{ssec:ihc_validation}) we contrasted Atlas H\&E-TME with H\&E-only pathologists at two fixed operating points: when every cell must be assigned a class (exhaustive annotation), and when the model may abstain on cells it cannot confidently call. 
The abstention setting is naturally a continuum rather than a single point: a model that exposes an informative confidence signal can be operated anywhere along a trade-off between \emph{coverage} (i.e., the fraction of cells that receive a prediction) and the accuracy of the cells it does predict. 
We analyze this trade-off in more detail here. 
 
\paragraph{Selective prediction setup.}
For each detected cell, Atlas H\&E-TME produces a posterior distribution $p$ over the cell classes. 
We use the predictive entropy $H(p) = -\sum_c p_c \log p_c$ as the model's uncertainty score and define a selective classifier that returns a label only when $H(p) \le \tau$ and abstains otherwise \citep{el2010foundations, geifman2017selective}. 
Sweeping the threshold $\tau$ generates the coverage-accuracy curve: at full coverage (100\%) the model labels every cell, while lower thresholds retain only the most confident cells. 
At each threshold we evaluate macro F1 over the five 5-plex classes on the cells the model retains at that threshold --- its own confident subset. 
The two in-depth operating points lie on this curve: its full-coverage limit corresponds to the exhaustive comparison
(Figure~\ref{fig:5plex_f1}), in which the model and the pathologists are scored on the same cells without abstaining, while the coverage-matched comparison (Figure~\ref{fig:5plex_f1_uncertain}) is the point at which the model's threshold is set to the average pathologist coverage and it is scored on the cells it keeps. 
 
\begin{figure}[tbh]
  \centering
  \includegraphics[width=0.75\textwidth]{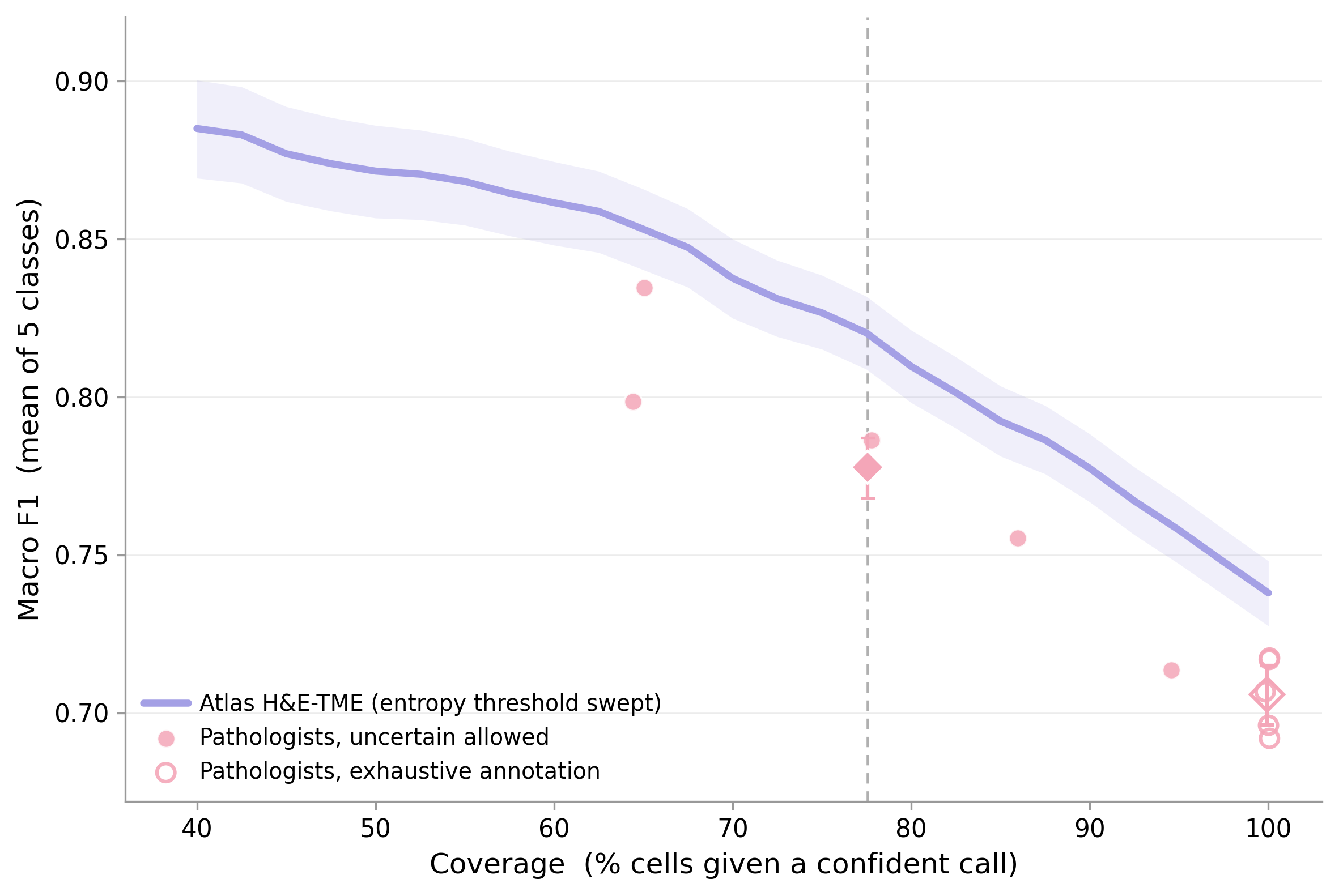}
  \caption{\textbf{Coverage--accuracy trade-off for Atlas H\&E-TME (macro F1).}
  Macro F1 (mean of the five 5-plex classes) as a function of coverage (percentage of cells
  given a confident call), obtained by sweeping the model's entropy threshold (purple line;
  shaded band, 95\% bootstrap CI). Pathologists are shown at their confidently labeled cells
  (filled circles) and when labeling every cell (open circles, coverage = 100\%);
  filled and open diamonds mark the respective pathologist means ($\pm\,$95\% CI).
  The dashed line indicates the average pathologist coverage ($\approx$ 78\%), the operating
  point of the coverage-matched comparison. Compared at its own
  coverage, the model curve lies at or above every individual pathologist operating point.}
  \label{fig:risk_coverage_macro}
\end{figure}

For reference we overlay each pathologist twice: once using the cells they confidently labeled (filled circles, at each pathologist's own coverage), and once when asked to label every cell (open circles, at 100\% coverage). 
Filled and open diamonds mark the corresponding pathologist means ($\pm\,$95\% CI). 
The dashed line in \Cref{fig:risk_coverage_macro} marks the average pathologist coverage ($\approx$ 78\%, i.e.\ a $\approx$ 22\% uncertainty rate), the coverage at which the model is compared to pathologists in Figure~\ref{fig:5plex_f1_uncertain}.

\paragraph{Aggregate behavior.}
\Cref{fig:risk_coverage_macro} shows the aggregate trade-off. 
Macro F1 increases steadily as coverage decreases, confirming that the entropy score is informative: the cells the model declines to call are disproportionately those it would otherwise misclassify. 
Compared at its own coverage, the Atlas H\&E-TME curve lies at or above every individual pathologist operating point. 
It exceeds the pathologist mean at the matched ($\approx$ 78\%) coverage, and remains above the pathologist mean when every cell is labeled, where the model has no abstention advantage. 
The matched-coverage result reported in the main text is therefore not an artifact of a particular threshold but holds along the whole continuum. 
 
\begin{figure}[tbh]
  \centering
  \includegraphics[width=0.85\textwidth]{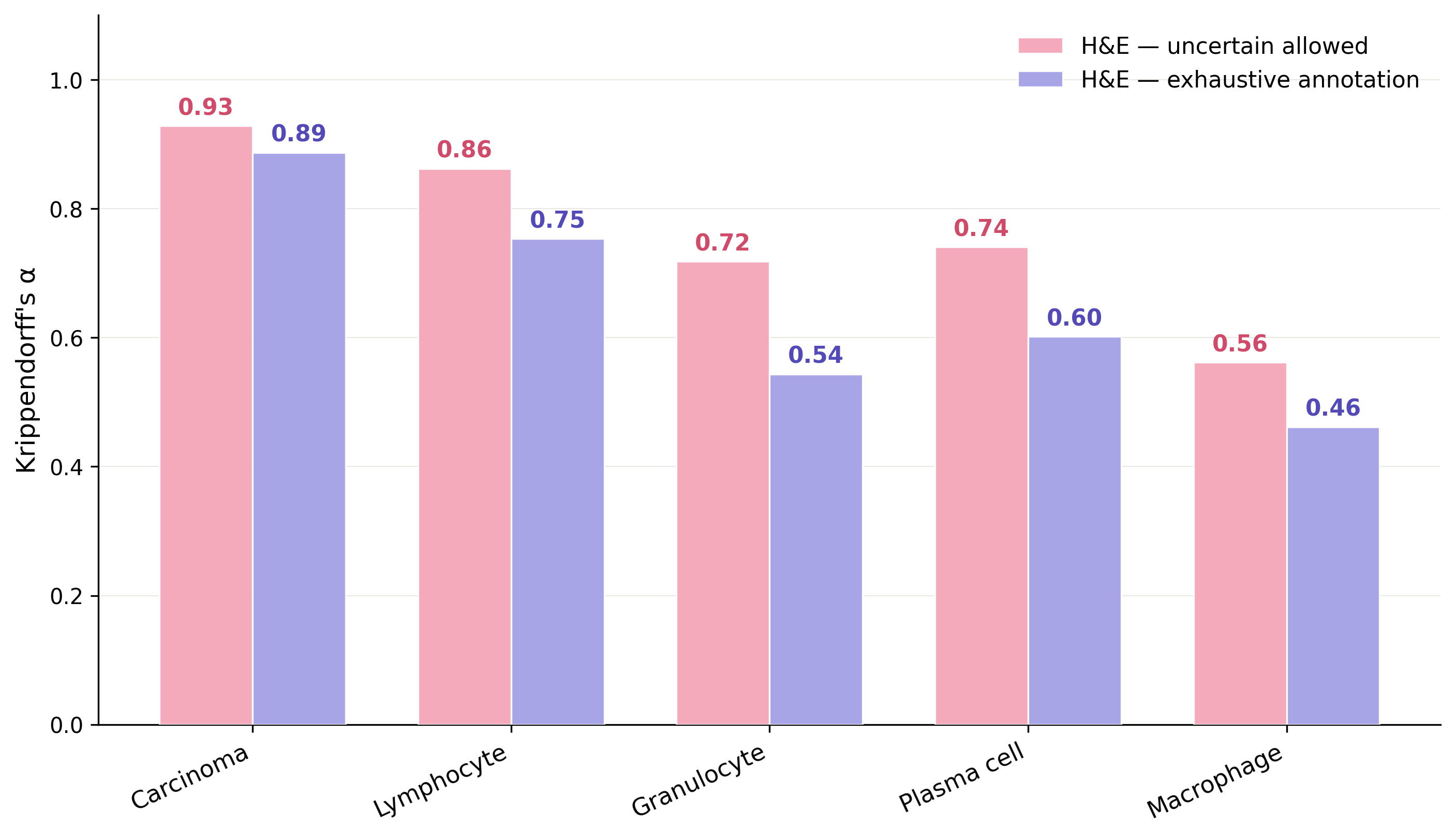}
  \caption{Per-class inter-rater agreement (Krippendorff's $\alpha$) among the
  five pathologists on H\&E only, contrasting the two annotation protocols:
  allowing an explicit \emph{uncertain} call (pink) versus forcing a decision on
  every cell (purple). The uncertain allowed bars coincide with the H\&E-only condition
  of Figure~\ref{fig:5plex_alpha} by construction.}
  \label{fig:5plex_alpha_he_uncertain}
\end{figure}

\paragraph{Effect of permitting abstention on inter-rater agreement.} To quantify how the annotation protocol itself shapes reproducibility, we recomputed H\&E-only inter-rater agreement under the two regimes among the same five pathologists (Figure~\ref{fig:5plex_alpha_he_uncertain}). 
When annotators are required to commit to a class on every cell, agreement drops markedly relative to when they may flag ambiguous cells as \emph{uncertain} (which we treat as missing data in Krippendorff's $\alpha$). 
Permitting abstention raises $\alpha$ for every cell type, from a mean of 0.65 (no uncertain) to 0.76 (uncertain allowed). 
The gain is smallest for the morphologically distinctive carcinoma cells (0.89 → 0.93) and largest for the classes that are hardest to call on H\&E alone --- granulocytes (0.54 → 0.72), plasma cells (0.60 → 0.74), and lymphocytes (0.75 → 0.86). 
This indicates that much of the disagreement observed under exhaustive annotation (i.e., no ``uncertain'' option) stems from raters being compelled to guess on low-confidence cells: once those cells are set aside, the remaining confident calls agree substantially more. 
The effect is only partial for macrophages, which remain the least reproducible class even with abstention (0.46 → 0.56), consistent with genuine residual ambiguity that an \emph{uncertain} option alone does not resolve. 
The strong agreement on confident calls --- particularly for carcinoma and lymphocytes ($\alpha \geq$ 0.86) --- confirms that restricting to high-confidence annotations yields a reliable signal, and motivates their use as the basis for the in-breadth analysis. 

 
\clearpage

\section{Additional In-Depth Validation Cohort}
\label{app:capri}

We present an additional in-depth validation cohort drawn from the NSCLC tissue microarray (TMA) dataset, results, and IHC-informed annotation protocol established in \cite{mrowiec2022immunohistochemistry} --- to our knowledge, the earliest application of a multi-pathologist IHC-informed consensus framework for benchmarking H\&E-based cell classification. 
We revisit this cohort here as a second, methodologically distinct in-depth validation point: updating the original three-class analysis to the current Atlas H\&E-TME model, and extending the cell-class scope to the same five immune and tumor populations evaluated in the main 5-plex in-depth cohort of Section~\ref{ssec:ihc_validation}.

\paragraph{Cohort and protocol.}
The cohort consists of FFPE NSCLC tissue microarrays from \cite{mrowiec2022immunohistochemistry}, sourced from Charité – Universitätsmedizin Berlin and University Hospital Cologne and scanned on a Roche Ventana DP200 scanner. 
Sections were stained with H\&E and scanned, then bleached and sequentially re-stained with single-marker IHC antibodies targeting individual cell populations: CK-KL1 for carcinoma cells, CD3+CD20 for lymphocytes, MUM1 for plasma cells, a CD68/MPO dual-chromogen stain distinguishing granulocytes (MPO-red) from macrophages (CD68-brown). 
H\&E and IHC scans were coregistered at micrometer precision and individual TMA cores were extracted as the unit of evaluation. 
Three board-certified pathologists then independently annotated each core in two passes separated by a washout period: a first pass on H\&E alone, followed by a second pass with the IHC-derived information available. In contrast to the main in-depth evaluation, in which pathologists assign labels to pre-detected cells, here each pathologist outlined every cell free-hand; the two-pass H\&E-then-IHC washout structure nonetheless follows that of the main in-depth evaluation.

An IHC-informed consensus reference was constructed by matching the three pathologists' second-pass annotations across raters by intersection-over-union (IoU) and assigning each matched cell a majority-vote label. Atlas H\&E-TME was compared against the H\&E-only pathologist mean by matching its detected cells to each pathologist's confidently labeled cells (their H\&E-only coverage) by IoU, scored symmetrically on each pathologist's coverage subset and averaged across raters. This cohort is therefore evaluated only in this coverage-restricted setting, not under the exhaustive primary comparison of Section~\ref{ssec:ihc_validation}.

\paragraph{Methodological differences from the main in-depth protocol.}
Beyond the free-hand, IoU-matched annotation protocol described above, several aspects of this cohort differ from the 5-plex multi-indication protocol of Section~\ref{ssec:ihc_validation} in ways that constrain the per-cell information available at annotation time and limit direct comparability of absolute scores across the two cohorts. 
\emph{(i)} IHC was applied as single-marker stains rather than as a 5-marker multiplex panel, so each IHC channel resolves a single cell class at a time and the IHC-informed reference is built from independent single-marker passes rather than a jointly coregistered multi-marker panel. 
\emph{(ii)} Tissue is presented as TMA cores rather than full resection sections, reducing the surrounding tissue context available to pathologists at annotation time and to Atlas H\&E-TME at inference time. 
\emph{(iii)} The IHC-informed consensus is constructed from three pathologists rather than five, providing a smaller annotator pool over which to average out individual rater idiosyncrasies in both the H\&E-only and IHC-informed annotation passes. 
These differences tighten the conditions under which pathologists annotate from H\&E alone more than the conditions under which Atlas H\&E-TME predicts from H\&E. 
We accordingly treat this cohort as a corroborating in-depth validation under similar but methodologically distinct and less elaborate conditions, not as a replacement for the main 5-plex in-depth evaluation. 

\begin{figure}[h]
    \centering
    \includegraphics[width=0.85\textwidth]{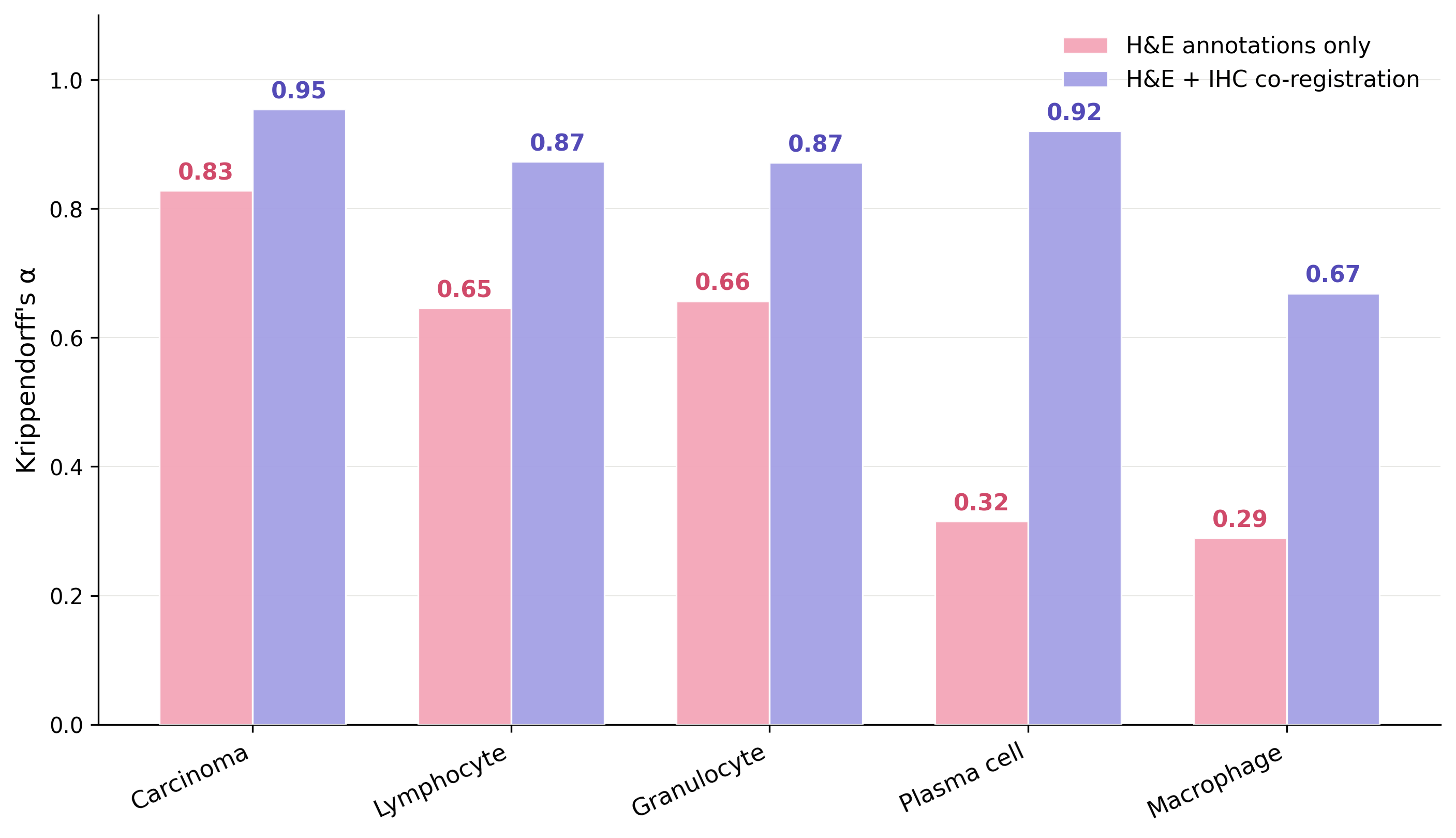}
    \caption{Inter-rater agreement (Krippendorff's $\alpha$) between the three pathologists on the NSCLC TMA cohort of \cite{mrowiec2022immunohistochemistry}, across the five evaluated cell classes, comparing H\&E-only annotation to IHC-informed annotation. Agreement increases consistently from H\&E-only to IHC-informed annotation, with the largest gains on the morphologically most ambiguous immune populations, mirroring the pattern observed on the main 5-plex cohort (Figure~\ref{fig:5plex_alpha}). Absolute H\&E-only $\alpha$ baselines are systematically lower than on the 5-plex cohort, explained by the methodological differences of this cohort (e.g., reduced per-cell tissue context of TMA cores, etc.).}
    \label{fig:capri_alpha}
\end{figure}

\paragraph{IHC-informed consensus improves inter-rater agreement.}
Krippendorff's $\alpha$ between the three pathologists increases consistently from H\&E-only to IHC-informed annotation across all five evaluated cell classes (Figure~\ref{fig:capri_alpha}). 
The gains are largest on the morphologically most ambiguous immune classes --- plasma cells ($0.32 \to 0.92$) and macrophages ($0.29 \to 0.67$) --- and remain substantial for lymphocytes ($0.65 \to 0.87$), granulocytes ($0.66 \to 0.87$), and carcinoma cells ($0.83 \to 0.95$). 
H\&E-only $\alpha$ baselines are systematically lower on this cohort than on the 5-plex cohort (cf. Figure~\ref{fig:5plex_alpha}), consistent with the reduced per-cell tissue context provided by TMA cores and the less standardized annotation conventions. 
IHC-informed $\alpha$ values reach a comparable regime to the 5-plex cohort for carcinoma, lymphocyte, granulocyte, and plasma cells, while remaining lower for macrophages, where the single-marker CD68 stain offers weaker disambiguation than the CD68/MPO and MPO/CD68 dual-chromogen design of the 5-plex panel (Table~\ref{tab:5plex_ihc_panel}). 
The qualitative finding --- that IHC-informed annotation establishes a more reliable reference than H\&E-only annotation, particularly for the morphologically ambiguous immune populations --- replicates the key finding of the main in-depth cohort. 

\begin{figure}[h]
    \centering
    \includegraphics[width=0.85\textwidth]{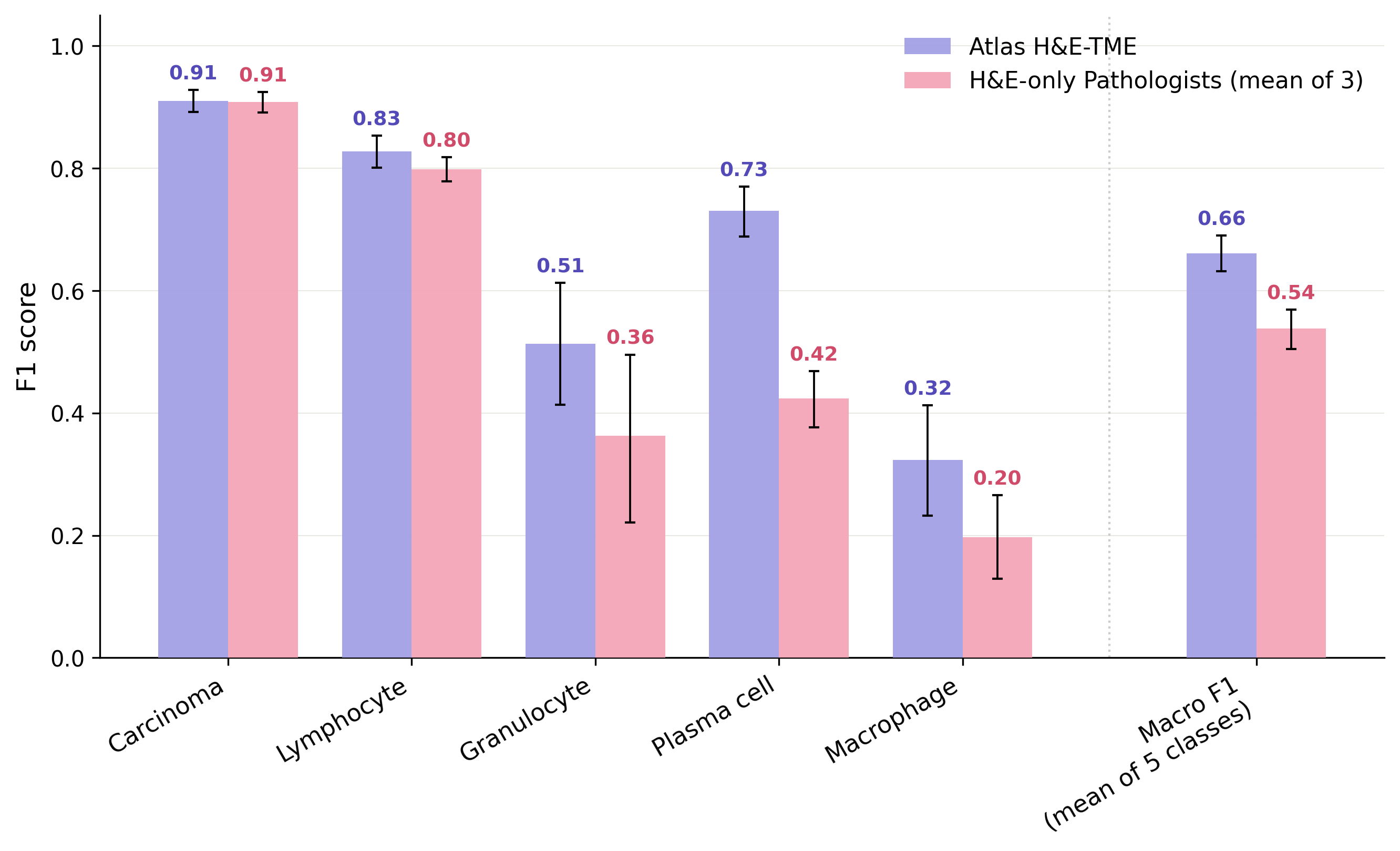}
    \caption{Per-class F1 score of Atlas H\&E-TME and the H\&E-only pathologist mean against the IHC-informed three-pathologist consensus on the NSCLC TMA cohort of \cite{mrowiec2022immunohistochemistry}. Atlas H\&E-TME matches or exceeds the H\&E-only pathologist mean across all five evaluated cell classes, with overlapping or favorable 95\% confidence intervals (10k stratified bootstrap iterations). Absolute F1 values are lower than on the main 5-plex cohort (Figure~\ref{fig:5plex_f1}) for both Atlas H\&E-TME and the pathologist mean, with pathologist scores affected more strongly, explained by the methodological differences of this cohort.}
    \label{fig:capri_f1}
\end{figure}

\paragraph{Atlas H\&E-TME matches or exceeds pathologist H\&E performance.}
Evaluated against the IHC-informed three-pathologist consensus, Atlas H\&E-TME matches or exceeds the H\&E-only pathologist mean F1 across all five evaluated cell classes (Figure~\ref{fig:capri_f1}), with overlapping or favorable 95\% confidence intervals throughout. 
Macro F1 across the five classes is 0.66 for Atlas H\&E-TME and 0.54 for the pathologist H\&E mean.
Absolute F1 values are lower on this cohort than on the 5-plex cohort for both Atlas H\&E-TME and the pathologist mean, with pathologist scores affected more strongly than Atlas H\&E-TME scores --- consistent with the methodological differences enumerated above acting more strongly on the H\&E-only annotation process than on H\&E-only model inference. 
Despite this overall shift, the qualitative finding of the main 5-plex cohort --- that Atlas H\&E-TME matches or exceeds H\&E-only pathologist performance when both are referenced against an IHC-informed multi-pathologist consensus --- holds on this independent cohort under a methodologically distinct in-depth protocol. 
The two in-depth cohorts thus converge on the same key conclusion across different design choices in cohort format, IHC multiplexing, pathologist pool, and annotation tooling.

\end{document}

%% file: tab/5plex_markers.tex
\begin{tabular}{lll}
    \toprule
    \textbf{IHC marker} & \textbf{Chromogen} & \textbf{Target cell class} \\
    \midrule
    Pan-CK                  & yellow & Carcinoma cell \\
    CD3 + CD20              & brown  & Lymphocyte \\
    MPO$^{+}$/CD68$^{-}$    & purple & Granulocyte \\
    MUM1                    & green  & Plasma cell \\
    CD68$^{+}$/MPO$^{-}$    & teal   & Macrophage \\
    \bottomrule
\end{tabular}

%% file: tab/in-depth_ranking.tex
\begin{tabular}{lccccccc}
\toprule
 & P1 & P2 & P3 & P4 & P5 & \multicolumn{2}{c}{Atlas H\&E-TME} \\
\cmidrule(lr){7-8}
 & & & & & & F1 & Rank \\
\midrule
Carcinoma cell & 0.900 & 0.907 & 0.907 & 0.908 & 0.920 & 0.921 & 1 \\
Lymphocyte     & 0.817 & 0.847 & 0.822 & 0.831 & 0.794 & 0.878 & 1 \\
Granulocyte    & 0.623 & 0.595 & 0.611 & 0.655 & 0.573 & 0.645 & 2 \\
Plasma cell    & 0.630 & 0.696 & 0.657 & 0.679 & 0.659 & 0.702 & 1 \\
Macrophage     & 0.515 & 0.543 & 0.461 & 0.515 & 0.588 & 0.542 & 3 \\
\midrule
\textbf{Macro F1} & \textbf{0.697} & \textbf{0.718} & \textbf{0.692} & \textbf{0.718} & \textbf{0.707} & \textbf{0.738} & \textbf{1} \\
\bottomrule
\end{tabular}

%% file: tab/in-breadth_primary.tex
\begin{tabular}{lcccccccccc}
\toprule
 & Bladder & Breast & Colorectal & Liver & \multicolumn{2}{c}{Lung} & Pancreas & Prostate & Stomach \\
\cmidrule(lr){6-7}
 & & & & & NSCLC & NET & & & \\
\midrule
Carcinoma cell      & 0.98 & 0.96 & 0.99 & 0.98 & 0.99 & 0.99 & 0.98 & 0.96 & 0.96 \\
Lymphocyte          & 0.97 & 0.95 & 0.95 & 0.96 & 0.95 & 0.88 & 0.98 & 0.87 & 0.92 \\
Plasma cell         & 0.93 & 0.87 & 0.96 & 0.91 & 0.92 & 0.94 & 0.89 & 0.74 & 0.90 \\
Epithelial cell     & 0.87 & 0.88 & 0.96 & 0.96 & 0.93 & 0.92 & 0.91 & 0.91 & 0.95 \\
Macrophage          & 0.74 & 0.78 & 0.71 & 0.81 & 0.94 & 0.73 & 0.72 & 0.75 & 0.69 \\
Granulocyte         & 0.97 & 0.84 & 0.93 & 0.94 & 0.93 & 0.90 & 0.94 & 0.85 & 0.96 \\
Endothelial cell    & 0.97 & 0.97 & 0.96 & 0.92 & 0.96 & 0.97 & 0.98 & 0.97 & 0.97 \\
Fibroblast          & 0.82 & 0.93 & 0.91 & 0.96 & 0.93 & 0.95 & 0.95 & 0.78 & 0.87 \\
Other               & 0.93 & 0.95 & 0.95 & 0.96 & 0.95 & 0.88 & 0.97 & 0.83 & 0.95 \\
\midrule
\textbf{Macro F1} & \textbf{0.91} & \textbf{0.90} & \textbf{0.92} & \textbf{0.93} & \textbf{0.94} & \textbf{0.91} & \textbf{0.92} & \textbf{0.85} & \textbf{0.91} \\
\bottomrule
\end{tabular}

%% file: tab/in-breadth_metastatic.tex
\begin{tabular}{lccccc}
\toprule
 & Bone & Brain & Liver & Lung & Lymph node \\
\midrule
Carcinoma cell      & 0.98 & 0.99 & 0.99 & 0.99 & 0.98 \\
Lymphocyte          & 0.90 & 0.91 & 0.84 & 0.92 & 0.93 \\
Plasma cell         & 0.65 & 0.92 & 0.74 & 0.94 & 0.89 \\
Epithelial cell     & --   & --   & 0.98 & 0.97 & 0.91 \\
Macrophage          & 0.54 & 0.78 & 0.65 & 0.97 & 0.79 \\
Granulocyte         & 0.83 & 0.81 & 0.87 & 0.95 & 0.91 \\
Endothelial cell    & 0.98 & 0.94 & 0.90 & 0.94 & 0.95 \\
Fibroblast          & 0.95 & 0.89 & 0.96 & 0.93 & 0.94 \\
Other               & 0.94 & 0.95 & 0.95 & 0.98 & 0.96 \\
\midrule
\textbf{Macro F1} & \textbf{0.85} & \textbf{0.90} & \textbf{0.88} & \textbf{0.95} & \textbf{0.92} \\
\bottomrule
\end{tabular}

%% file: tab/classes_tissue_qc.tex
\begin{tabular}{@{}lp{10cm}@{}}
\toprule
\textbf{Region} & \textbf{Description} \\
\midrule
Valid Tissue &
  Well-preserved and clearly identifiable tissue suitable for downstream analysis. In the subsequent steps only regions identified as Valid Tissue are processed. \\  
Out-of-Focus &
  Blurry or unclear region due to imaging issues. \\
Tissue Artifact &
  Distorted or altered tissue caused by staining, processing, or technical errors. \\
Marker &
  Physical drawings that a pathologist made on the glass slide with a pen. \\
\bottomrule
\end{tabular}

%% file: tab/classes_tissue_segmentation.tex
\begin{tabular}{@{}lp{10cm}@{}}
\toprule
\textbf{Tissue Type} & \textbf{Description} \\
\midrule
Carcinoma &
  Regions containing carcinoma and immune cells. In case of squamous cell carcinoma this category also includes intra- and extracellular keratin stemming from squamous carcinoma; dysplastic serous epithelium featuring high-grade architectural and/or cytological atypia but without proven invasive growth. \\
Epithelial Tissue &
  Regions containing epithelial and immune cells. Includes: mesothelial lining of peritoneal surfaces; epithelial cell layer with ciliated/squamous epithelium and enclosed immune cells; serous glandular cells including lumen, inner epithelial layer, outer basal layer and enclosed immune cells; mucinous glandular cells other than Brunner glands including lumen, inner epithelial layer, outer basal layer and enclosed immune cells; ductal epithelium including lumen, inner epithelial layer, outer basal layer and enclosed immune cells. \\
Stroma &
  Regions containing connective tissue and immune cells. Also includes secondary and tertiary lymphoid structures. \\
Necrosis &
  Regions of dying or dead cells. Includes coagulative, liquefactive, caseous, or fat necrosis with no discernible cell nuclei as well as
  partly discernible, potentially fragmented nuclei and enclosed immune cells. \\
Blood &
  Regions with accumulation of erythrocytes and/or fibrin, as seen in vessel lumina or hemorrhage. \\
Vessel &
  All vascular structures, including arteries, veins, arterioles, venules, capillary-like vessels ($\leq$10~\textmu m, $\leq$5
  erythrocytes, single endothelial layer, no tunica media/adventitia), and lymphatic vessels. Includes arteries with round lumens, thick muscular walls, and visible elastica; veins with irregular lumens, thinner muscular walls, and no visible elastica; lymphatics with irregular lumens, thin walls, and lymphatic fluid. \\
Other &
  Any tissue that does not fit into the predefined tissue types above. This includes tissue types such as mucus, fat, smooth muscle, bone, and anthracosis. \emph{Note:} In lung specimens, alveoli are typically classified under the ``Other'' tissue type; however, their constituent cells may be individually labeled as epithelial cells. \\
\bottomrule
\end{tabular}

%% file: tab/classes_cell_classification.tex
\begin{tabular}{@{}lp{10cm}@{}}
\toprule
\textbf{Cell Type} & \textbf{Description} \\
\midrule
Carcinoma cell &
  Malignant epithelial cell with uncontrolled growth and potential for invasion. Includes: carcinoma with glandular differentiation; carcinoma with squamous differentiation; carcinoma with squamous differentiation and intra- or extracellular keratin accumulation; non-small cell tumor with neuroendocrine differentiation; signet-ring cell carcinoma with cytoplasmic mucin and an eccentric, compressed nucleus; mucinous adenocarcinoma; carcinoma with cytoplasmic clearing; atypical serous cells with high-grade dysplasia without proven invasive growth; and atypical squamous cells with high-grade dysplasia without proven invasive growth, with possible keratinization. \\
Epithelial cell &
  Structural cells forming protective layers on organs and internal surfaces as well as comprising parenchyma of most organs. Includes: nonspecific benign epithelial cells; flat squamous-type cells, with or without intra-/extracellular keratin; columnar polarized cells with hair-like projections; mucin-containing polarized cells; basal cuboidal cells at the basement membrane; glandular pyramid-shaped cells with basally placed nuclei and zymogen granules; glandular cells with oval nuclei and mucin granules; small hormone-releasing cells with round/oval nuclei and salt-and-pepper chromatin; thin spindle-shaped cells adjacent to glandular epithelium; variably shaped duct-lining cells; and elongated flattened cells lining pleural and peritoneal surfaces. \\
Fibroblast &
  Connective tissue cell responsible for collagen production and extracellular matrix maintenance. \\
Lymphocyte &
  Small immune cells involved in adaptive immunity, including T and B cell responses. Includes: immune cells with small round nuclei and inconspicuous cytoplasm and activated lymphocytes (centroblasts and centrocytes) within the germinal center with large, cleaved or vesicular nuclei, nucleoli, and distinguishable cytoplasm. \\
Plasma cell &
  Differentiated B cell specialized in antibody production for adaptive immune defense. \\
Macrophage &
  Large immune cell of innate immune response that engulfs pathogens and debris while regulating immune responses. \\
Granulocyte &
  White blood cell with cytoplasmic granules, including neutrophils, eosinophils, basophils, and mast cells. \\
Endothelial cell &
  Specialized cell lining blood vessels, regulating permeability and vascular function. \\
Other &
  Any cell type that does not fit the predefined categories. \\
\bottomrule
\end{tabular}

%% file: tab/in-breadth_primary_ts.tex
\begin{tabular}{lccccccccc}
\toprule
 & Bladder & Breast & Colorectal & Liver & \multicolumn{2}{c}{Lung} & Pancreas & Prostate & Stomach \\
\cmidrule(lr){6-7}
 &         &        &            &       & NSCLC & NET                &          &          &         \\
\midrule
Carcinoma & 0.97 & 0.98 & 0.98 & 0.98 & 0.98 & 0.99 & 0.94 & 0.95 & 0.95 \\
Epithelial tissue & 0.93 & 0.96 & 0.94 & 0.98 & 0.92 & 0.98 & 0.90 & 0.95 & 0.93 \\
Stroma & 0.86 & 0.96 & 0.94 & 0.96 & 0.95 & 0.96 & 0.97 & 0.88 & 0.92 \\
Necrosis & 0.96 & 0.89 & 0.95 & 0.94 & 0.93 & 0.94 & 0.95 & 0.93 & 0.89 \\
Blood & 0.98 & 0.98 & 0.90 & 0.88 & 0.95 & 0.92 & 0.94 & 0.95 & 0.98 \\
Vessel & 0.96 & 0.82 & 0.92 & 0.89 & 0.89 & 0.94 & 0.96 & 0.97 & 0.91 \\
Other & 0.93 & 0.99 & 0.91 & 0.97 & 0.95 & 0.97 & 0.98 & 0.84 & 0.94 \\
\midrule
\textbf{Macro F1} & \textbf{0.94} & \textbf{0.94} & \textbf{0.93} & \textbf{0.94} & \textbf{0.94} & \textbf{0.96} & \textbf{0.95} & \textbf{0.92} & \textbf{0.93} \\
\bottomrule
\end{tabular}

%% file: tab/in-breadth_metastatic_ts.tex
\begin{tabular}{lccccc}
\toprule
 & Bone & Brain & Liver & Lung & Lymph node \\
\midrule
Carcinoma & 0.98 & 0.98 & 0.97 & 0.98 & 0.98 \\
Epithelial tissue & -- & -- & 0.99 & 0.97 & 0.94 \\
Stroma & 0.93 & 0.86 & 0.93 & 0.94 & 0.98 \\
Necrosis & 0.92 & 0.98 & 0.99 & 0.92 & 0.92 \\
Blood & 0.95 & 0.95 & 0.96 & 0.94 & 0.96 \\
Vessel & 0.96 & 0.75 & 0.80 & 0.96 & 0.91 \\
Other & 0.98 & 0.98 & 0.97 & 0.95 & 0.99 \\
\midrule
\textbf{Macro F1} & \textbf{0.95} & \textbf{0.92} & \textbf{0.94} & \textbf{0.95} & \textbf{0.95} \\
\bottomrule
\end{tabular}

%% file: tab/in-breadth_primary_qc.tex
\begin{tabular}{lccccccccc}
\toprule
 & Bladder & Breast & Colorectal & Liver & \multicolumn{2}{c}{Lung} & Pancreas & Prostate & Stomach \\
\cmidrule(lr){6-7}
 &         &        &            &       & NSCLC & NET                &          &          &         \\
\midrule
Valid tissue & 0.99 & 0.99 & 1.00 & 1.00 & 0.99 & 0.98 & 0.98 & 0.99 & 0.99 \\
Out-of-focus & 0.77 & 0.74 & 0.94 & 0.91 & 0.77 & 0.91 & 0.75 & 0.78 & 0.86 \\
Tissue artifact & 0.90 & 0.91 & 0.76 & 0.97 & 0.94 & 0.88 & 0.88 & 0.90 & 0.92 \\
Marker & 0.89 & 0.91 & 0.99 & 0.97 & 0.98 & 0.96 & 0.89 & 0.79 & 0.95 \\
No tissue & 0.99 & 0.99 & 0.96 & 1.00 & 0.99 & 1.00 & 0.97 & 0.99 & 1.00 \\
\midrule
\textbf{Macro F1} & \textbf{0.91} & \textbf{0.91} & \textbf{0.93} & \textbf{0.96} & \textbf{0.93} & \textbf{0.94} & \textbf{0.89} & \textbf{0.89} & \textbf{0.94} \\
\bottomrule
\end{tabular}

%% file: tab/in-breadth_metastatic_qc.tex
\begin{tabular}{lccccc}
\toprule
 & Bone & Brain & Liver & Lung & Lymph node \\
\midrule
Valid tissue & 0.99 & 0.99 & 1.00 & 0.98 & 0.98 \\
Out-of-focus & 0.85 & 0.89 & 0.95 & 0.81 & 0.90 \\
Tissue artifact & 0.89 & 0.95 & 0.97 & 0.88 & 0.81 \\
Marker & 0.96 & 0.96 & -- & 0.97 & 0.96 \\
No tissue & 0.98 & 1.00 & 1.00 & 0.99 & 0.99 \\
\midrule
\textbf{Macro F1} & \textbf{0.94} & \textbf{0.95} & \textbf{0.98} & \textbf{0.92} & \textbf{0.93} \\
\bottomrule
\end{tabular}